\def\year{2021}\relax
\documentclass[letterpaper]{article} % DO NOT CHANGE THIS
\usepackage{aaai21}  % DO NOT CHANGE THIS
\usepackage{times}  % DO NOT CHANGE THIS
\usepackage{helvet} % DO NOT CHANGE THIS
\usepackage{courier}  % DO NOT CHANGE THIS
\usepackage[hyphens]{url}  % DO NOT CHANGE THIS
\usepackage[pdftex]{graphicx}
\usepackage{natbib}  % DO NOT CHANGE THIS AND DO NOT ADD ANY OPTIONS TO IT
\usepackage{caption} % DO NOT CHANGE THIS AND DO NOT ADD ANY OPTIONS TO IT
\usepackage{subcaption}
\usepackage{multirow}
\usepackage{multicol}
\usepackage{wrapfig}
\usepackage[switch]{lineno}
\usepackage{appendix}

\usepackage{booktabs}       % professional-quality tables
\title{NASTransfer: Analyzing Architecture Transferability in Large Scale Neural Architecture Search}
\author{Rameswar Panda$^{1,2}$, Michele Merler$^{1}$, Mayoore Jaiswal$^{3}$, Hui Wu$^{1,2}$, \\ Kandan Ramakrishnan$^{4}$, Ulrich Finkler$^{1}$, Chun-Fu Chen$^{1,2}$, Minsik Cho$^{1}\footnote{Currently at Apple}$, \\ Rogerio Feris$^{1,2}$, David Kung$^{1}$, Bishwaranjan Bhattacharjee$^{1}$ \\
}
\affiliations{$^{1}$IBM Research, $^{2}$MIT-IBM Watson AI Lab, $^{3}$NVIDIA, $^{4}$Baylor College of Medicine}

\begin{document}
%\linenumbers

\maketitle

\begin{abstract}
Neural Architecture Search (NAS) is an open and challenging problem in machine learning. While NAS offers great promise, the prohibitive computational demand of most of the existing NAS methods makes it difficult to directly search the architectures on large-scale tasks. The typical way of conducting large scale NAS is to search for an architectural building block on a small dataset (either using a proxy set from the large dataset or a completely different small scale dataset) and then transfer the block to a larger dataset. Despite a number of recent results that show the promise of transfer from proxy datasets, a comprehensive evaluation of different NAS methods studying the impact of different source datasets has not yet been addressed. In this work, we propose to analyze the architecture transferability of different NAS methods by performing a series of experiments on large scale benchmarks such as ImageNet1K and ImageNet22K. We find that: (i) The size and domain of the  proxy set  does not seem to influence architecture performance on the target dataset. On average, transfer performance of architectures searched using completely different small datasets (e.g., CIFAR10) perform similarly to the architectures searched directly on proxy target datasets. However, design of proxy sets has considerable impact on rankings of different NAS methods. (ii) While different NAS methods show similar performance on a source dataset (e.g., CIFAR10), they significantly differ on the transfer performance to a large dataset (e.g., ImageNet1K). (iii) Even on large datasets, random sampling baseline is very competitive, but the choice of the appropriate combination of proxy set and search strategy can provide significant improvement over it. We believe that our extensive empirical analysis will prove useful for future design of NAS algorithms.
\end{abstract}

%%%%%%%%% Introduction %%%%%%%%%
\section{Introduction}
\label{sec:introduction}

Neural Architecture Search (NAS) is a very active area of research \cite{survey_JMLR2019}, aiming at automatic design of deep learning networks for various applications spanning from image classification \cite{progressiveNAS_ECCV2018, MnasNet_CVPR2019, wu2018fbnet, randomly_wired_ICCV2019} to NLP \cite{liu2018darts, pmlr-v80-pham18a, NAS_RL_ICLR2017}, from object detection \cite{DetNAS_Neurips2019, NASFPN_CVPR2019, wang2019nasfcos} to semantic segmentation \cite{liu2019autodeeplab}.
%and control tasks \cite{weight_agnostic_neurips2019}. 
A number of NAS strategies have been proposed, including evolutionary methods \cite{baker2018accelerating, liu2018hierarchical, regularized_AAAI2019, EcoNAS_CVPR2020}, reinforcement learning \cite{progressiveNAS_ECCV2018, pmlr-v80-pham18a, ZhongYWSL18, NAS_RL_ICLR2017}, and gradient-based methods \cite{DATA_neurips2019, liu2018darts, nayman2019xnas, xie2019snas, Zela2020Understanding}. Despite impressive results on common benchmark datasets, the prohibitive computational demand of existing NAS methods makes it difficult to directly search the architectures on large-scale datasets (e.g., ImageNet). Motivated by this, many methods have been proposed to improve the efficiency of NAS by shifting the training and evaluation of candidate architectures from the entire target set to proxy tasks, which could mean learning with only a few blocks, training for a few epochs \cite{reduced_training_NCA2020} or using proxy sets \cite{liu2018darts, pmlr-v80-pham18a, EcoNAS_CVPR2020}. Proxy sets could either be smaller versions of the target dataset obtained through sampling, or datasets with similar distribution to the target, but with reduced number of classes or number of examples per class. 

Despite a number of recent work showing promising transfer results, comparison between different NAS methods in terms of architecture transfer remains a novel and rarely addressed problem. Specifically, it is not clear to what extent the architectures depend on the proxy dataset on which the search is conducted and how does the performance on the target dataset depend on the NAS method that is used to search the architectures. Moreover, a thorough study on applicability of proxy datasets to large scale contexts such as ImageNet22K is still missing. In fact, even direct training and evaluation of standard human-designed architectures has been relatively limited for ImageNet22K, given not only its sheer scale ($\sim$14M images) but also its large imbalance across classes \cite{cho2017powerai, adamOSDI2014, intel_blog_2017, zhang2015poseidon}. 

Motivated by this, in this paper, instead of focusing on beating the latest benchmark numbers on small scale datasets like CIFAR10~\cite{Krizhevsky09learningmultiple}, we take a step back and aim at filling the above gap with an extensive empirical study on architecture transferability of different NAS methods that explains and suggests best practices for proxy sets design and successful transfer at large scale. We compare four representative NAS methods such as ENAS~\cite{pmlr-v80-pham18a}, NSGANet~\cite{NSGA_Net_2019}, NAO~\cite{NAO} and DARTS~\cite{liu2018darts} using the commonly used DARTS search space on six diverse datasets, MIT67~\cite{quattoni2009recognizing}, FLOWERS102~\cite{nilsback2008automated}, CIFAR10~\cite{Krizhevsky09learningmultiple}, CIFAR100~\cite{Krizhevsky09learningmultiple}, ImageNet1K~\cite{imagenet_cvpr09} and ImageNet22K~\cite{ILSVRC15}, to analyze their transfer performance under different settings. Our findings suggest that transfer performance of architectures searched using completely different small datasets perform similarly to the architectures searched directly on proxy target datasets (e.g., CIFAR10 proves to be a valuable dataset for transferring architectures to ImageNet22K). Reliably good  search  for  large  scale  datasets  can  be  performed on proxy sets even smaller than two orders of magnitude  with  respect  to  the  target  ones.

Furthermore, we show that (a) While different NAS methods show similar performance on a source dataset, they significantly differ on the transfer performance to a large dataset. (b) Even on large datasets, random sampling remains a strong baseline to surpass, but the choice of the appropriate combination of proxy set and search strategy can provide significant improvement over it.

%%%%%%%%% Related Work %%%%%%%%%
\section{Related Work}
\label{sec:relatedwork}

\textbf{Neural Architecture Search.} % \label{ssec:rw_nas}
Neural Architecture Search has attracted intense attention in recent years. 
Typically, a NAS algorithm first defines a search space and then employs a search strategy within that space. During the search phase, some evaluation criteria are chosen to rank the relative performance of possible architecture candidates \cite{survey_JMLR2019, Yu2020Evaluating}. Recent studies \cite{Dong2020NAS-Bench-201, LiTUAI19, Yang2020NAS, pmlr-v97-ying19a} have shown that performance is highly dependent on the elaborately designed search space, within which the difference between different search strategies results less significant than initially thought, especially when compared to random search \cite{LiTUAI19, Yang2020NAS}. 
On the other hand, the search phase for candidate architectures within the search space highly influences the efficiency of a NAS algorithm. 
The original reinforcement learning based method \cite{NAS_RL_ICLR2017}, for example, required hundreds of GPUs in order to evaluate and rank each proposed architecture.
Different methods have been proposed to reduce the search and evaluation costs, including  micro-search of primary building cells \cite{ZhongYWSL18, ZophVSL17}, prediction of candidate architectures performance based on learning curves \cite{baker2018accelerating,finkler2020large} or surrogate models \cite{liu2018darts}, and parameter sharing between child models \cite{pmlr-v80-bender18a, brock2018smash, liu2018darts, pmlr-v80-pham18a, Zela2020NAS-Bench-1Shot1}. 

\vspace{1mm}
\noindent\textbf{Architecture Transferability.} 
Most NAS approaches usually perform well when searching an architecture for a specific dataset and/or task, but have a hard time generalizing.  In order to overcome the computational burden of running NAS searches for every new target domain, methods have been developed for joint training and efficient transfer of prior knowledge between multiple search spaces and tasks \cite{Borsos2019TransferNK, wistuba2019xfernas, lu2020neural}. Some methods obtain transferability based on meta-learning \cite{Lian2020Towards} or learning general supernets from which specialized subnets can be sampled without any additional training \cite{lu2020neural}. Other approaches search for the best cell on a small proxy dataset and
then trasnfer to the large target dataset by stacking together more copies of this cell, each with their own parameters to design a convolutional architecture \cite{ZophVSL17}. In this work, we focus on the latter type of approaches and investigate their applicability at large scale.

\vspace{1mm}
\noindent\textbf{NAS Proxies.} %\label{ssec:rw_proxies}
Although recent NAS methods \cite{progressiveNAS_ECCV2018, liu2018darts, pmlr-v80-pham18a} improve the search efficiency to some extent, the search process is still time-consuming and requires vast computation overhead when searching in a large search space since all network candidates need to be trained and evaluated. Differentiable approaches such as DARTS \cite{liu2018darts} require high GPU memory consumption, which still makes direct search on large dataset prohibitive.
A widely used approach to address efficiency in NAS methods is to search for an architectural building block on a small dataset (either using a proxy set from the large dataset or a completely different small scale dataset) and then transfer the block to the larger dataset by replicating and stacking it multiple times in order to increase network capacity according to the scale of the dataset. 
While proxy sets have been largely used to expand search results from small (CIFAR10, CIFAR100) to mid-size (ImageNet1K) datasets, and some works have been able to perform search on mid size datasets \cite{cai2019proxylessnas}, a study of their applicability to large scale datasets is still missing. 
In this work, we offer a detailed and extensive study of the effects of proxy sets on network transferability to large scale targets. We hope this will contribute to an established protocol of reproducibility when studying NAS algorithms going from small-to-medium proxies to large scale target datasets.

%%%%%%%%% Proposed Method %%%%%%%%%
\section{NASTransfer Benchmark}
\label{sec:proposedmethod}
In this section, we discuss the details about our proposed NASTransfer benchmark, in terms of datasets, methods and evaluation metric used to compare different methods. 

\vspace{1mm}
\noindent\textbf{Objective.} 
Our goal is to provide diagnostic information on the architecture transfer performance of different NAS methods and proxy sets for large scale NAS, which can be potentially used for better designs of future NAS algorithms. We adopt a common search space and training protocol to avoid the effect of the manually engineered tricks and search space widely used in different NAS methods.  

\vspace{1mm}
\noindent\textbf{Datasets and Proxy Sets.} 
We select six diverse and challenging computer vision datasets in image
classification, namely MIT67~\cite{quattoni2009recognizing}, FLOWERS102~\cite{nilsback2008automated}, CIFAR10 and CIFAR100~\cite{Krizhevsky09learningmultiple}, ImageNet1K~\cite{imagenet_cvpr09} and ImageNet22K~\cite{ILSVRC15} to evaluate the performance of different methods. While most of the existing analysis on NAS~\cite{Yang2020NAS,Zela2020NAS-Bench-1Shot1,pmlr-v97-ying19a,Dong2020NAS-Bench-201} focus on small scale datasets such as CIFAR10, we show large-scale experiments on the ImageNet22K~\cite{ILSVRC15} dataset that contains over 14 million labeled high-resolution images belonging to around 22K different categories. The ImageNet22K dataset skew is reflective of real world tasks and provides a natural testbed for our method when comparing training sets of different sizes. 

As proxy sets for the larger datasets, we employed not only small-scale datasets such as CIFAR10, but also sampled subsets of ImageNet1K and ImageNet22K directly. Namely, we investigated the proxies listed in Table \ref{tab:proxies}, which are of two types: randomly selected and uniformly selected. For random selection, we picked a list of $N$ classes and used all of their images. In uniform selection we were interested in maintaining the overall distribution of examples for all classes in the dataset, therefore we sorted classes by their number of images and then uniformly sampled in order to obtain the desired number of classes $N$ in the subset. In order to maintain the order of magnitude consistent across multiple proxies, we then took the same fraction of images from every class, ensuring that the total would meet the requirement and at the same time maintain the overall distribution intact. This is particularly important when designing a proxy set for a non-uniform, imbalanced distribution such as the one of ImageNet22K. For example ImageNet22K Proxy 2 was designed to have the same overall distribution of the full dataset, but the same number of images of ImageNet22K Proxy 1. In order to do so, we sampled 0.97\% of images from each class in the dataset, and eliminated classes for which only one image remained for training or validation, thus keeping only approximately 13k classes out the the 22k total. For ImageNet22K Proxy 3 instead we uniformly picked 100 classes whose total number of images would be the same as ImageNet22K Proxy 1.
We split each of those datasets into a training, validation and testing subsets with proportions 40/40/20 and use standard data pre-processing and augmentation techniques.

\vspace{1mm}
\noindent\textbf{Methods and Search Space.} 
We compare four representative NAS methods: DARTS~\cite{liu2018darts}, ENAS~\cite{pmlr-v80-pham18a}, NSGANet~\cite{NSGA_Net_2019} and NAO~\cite{NAO}, including the random sampling~\cite{Yang2020NAS} baseline. We choose these methods as they have a reasonable search time, specifically under 4 GPU-days on CIFAR10 dataset. 
We perform micro-search at cell level within the DARTS search \cite{liu2018darts}: 3$\times$3 and 5$\times$5 separable convolutions, 3$\times$3 and 5$\times$5 dilated separable convolutions, 3$\times$3 max pooling, 3$\times$3 average pooling, identity, and zero. All operations are of stride one (if applicable) and the convolved feature maps are padded to preserve their spatial resolution. ReLU-Conv-BN order are used for convolutional operations, and each separable convolution is always applied twice. Note that our NASTransfer benchmark has a fixed search space and hence provides a unified benchmark for analyzing transferability of different NAS algorithms.

\vspace{1mm}
\noindent\textbf{Training and Evaluation Protocol.}
NAS algorithms traditionally work in two phases: first \textit{search}, in which the best architecture is determined based on the search algorithm employed, and second \textit{augmentation}, which consists in training from scratch the model selected during the search phase. We choose the search hyperparameters as close as possible to the ones reported in the original papers. Experiments on all datasets use the same hyperparameters except the number of training epochs. For augmentations, we use cross entropy loss, SGD optimizer with learning rate 0.025, momentum 0.9, seed 2, initial number of channels 36, and gradient clipping set at 5. 
The impact of the seed was investigated in one ablation study for all methods in Table \ref{tab:aug_seed}.
While different augmentation strategies haven shown to be effective in improving the results, we did not use any such data augmentation strategy or other learning tricks to make a fair comparison among different NAS methods. We provide the effect of different augmentation strategies such as Drop path \cite{larsson2017fractalnet}, Auxiliary towers \cite{randomly_wired_ICCV2019} and Cutout \cite{devries2017cutout} in the supplementary material which shows that these widely used augmentation strategies have a larger impact on small datasets, but fails to provide consistent improvements on large datasets. 
The number of cells was fixed to 20 for all experiments and the number of training epochs per dataset was set to 600, 600, 600, 600, 120 and 60 for augment runs on MIT67, FLOWERS102, CIFAR10, CIFAR100, ImageNet1K and ImageNet22K, respectively.
%except the ablation study reported in Table \ref{tab:aug_cell}
All searches were performed on a single GPU, while augment runs were done on single GPU for CIFAR10 and CIFAR100. We use a minimum of 8 to a maximum of 96 GPUs for ImageNet1K and ImageNet22K augmentation experiments.

\iffalse
In all our experiments we used the codebase\footnote{\url{http://www.mediafire.com/file/ilqp3nqtr0l269e/NAS-Benchmark-master.zip}} from Yang et al. \cite{Yang2020NAS} running Pytorch on NVIDIA Tesla V100 GPUs with 16GB memory. 
\colorbox{Yellow}{Search details of parameters for each method.}
\colorbox{Yellow}{DARTS: Adam optimizer, alphas =, $alpha_lr$=, $alpha_{weight decay}$=}
\colorbox{Yellow}{betas=, lr =, seed =, weight decay=, momentum=, batch size=}
\colorbox{Yellow}{ENAS:}
\colorbox{Yellow}{NSGA-Net:}
\colorbox{Yellow}{NAO:}
Augment runs. 
For augmentattions, batch size was fixed at 128 per GPU across all augment runs, except for models using 40 and 60 cells (see Table \ref{tab:aug_cell}), where the memory constraints required smaller batch sizes of 96 and 64 per GPU, respectively.
For augmentations we followed the default settings as the experiments in \cite{Yang2020NAS}, using cross entropy loss, SGD optimizer with learning rate 0.025, momentum 0.9, seed 2, initial number of channels 36, and gradient clipping set at 5. The impact of the seed was investigated in one ablation study for all methods in Table \ref{tab:aug_seed}. 
\fi

\begin{table}[t]
\centering
\caption{\textbf{Proxy sets used in our experiments.} In Uniform selection the distribution of examples is reflective, although in smaller scale, of the overall distribution of the entire dataset. This is of particular relevance for ImageNet22K, where images are not uniformly distributed over classes. } 
\resizebox{0.47\textwidth}{!}{%
\begin{tabular}{c|c|c|c|c}
\toprule
Set   & Selection  & Number of   & Number of  & Number of  \\
Name & Method & Classes & Train. Images & Val. Images\\
\midrule
MIT67            & All & 67 & 5K  & 1.4K \\ 
FLOWERS102       & All & 102 & 6.5K  & 1.7K \\ 
CIFAR10             & All & 10 & 50K & 10K \\ 
CIFAR100            & All & 100 & 50K & 10K \\ 
ImageNet1K Proxy 1  & Random & 100 & 128K & 5K   \\ 
ImageNet1K Proxy 2  & Random & 200 & 258K  & 10K  \\ 
ImageNet1K Proxy 3  & Random & 300 & 384K & 15K   \\ 
ImageNet1K Proxy 4  & Random  & 200 & 128K  & 5K  \\ 
ImageNet1K Proxy 5  & Uniform & 1,000 & 128K & 5K  \\ 
ImageNet22K Proxy 1  & Random & 100 & 35K  & 35K \\ 
ImageNet22K Proxy 2  & Uniform & 13,377 & 35K & 35K  \\ 
ImageNet22K Proxy 3  & Uniform & 100 & 35K & 35K  \\ \hline
ImageNet1K   & All & 1,000 & 1.2M & 50K   \\ 
ImageNet22K   & All & 21,841 & 7.5M & 7M   \\ 
\bottomrule
\end{tabular}
}
\label{tab:proxies} \vspace{-10pt}
\end{table} 

\begin{figure*}[ht]
%\begin{minipage}{.6\textwidth}
  \centering
  %\captionsetup[subfigure]{labelformat=empty}
    \begin{subfigure}{0.4\linewidth}
        \includegraphics[width=\linewidth]{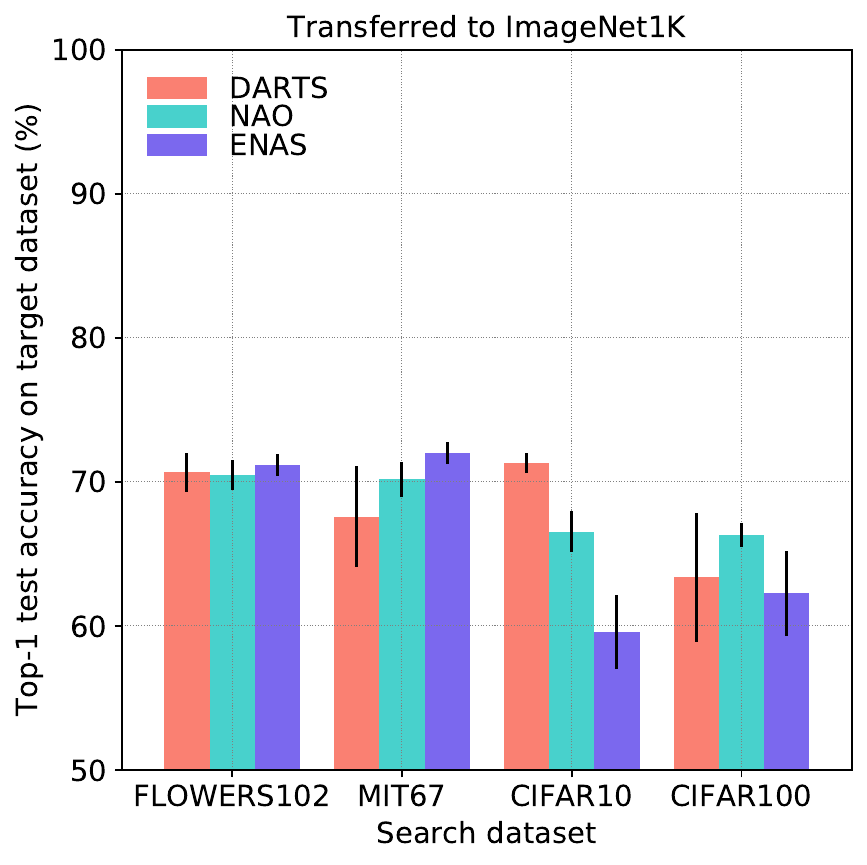}
    \end{subfigure}%
    \begin{subfigure}{0.4\linewidth}
    \includegraphics[width=\linewidth]{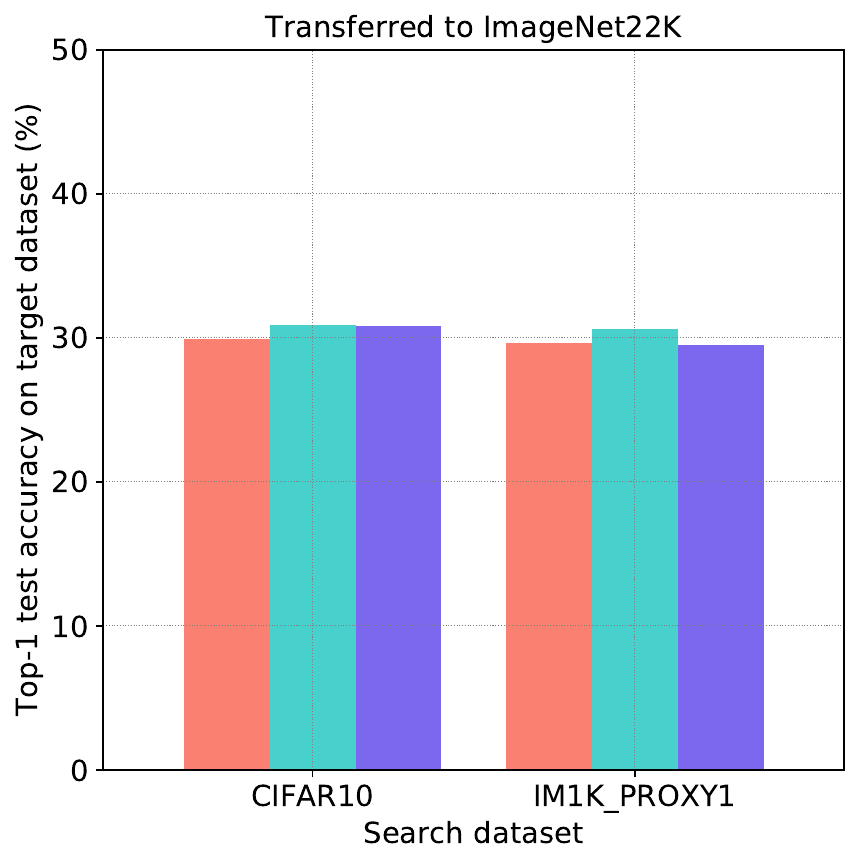}
    \end{subfigure} \vspace{-2mm}
  \caption{\textbf{Architecture transfer performance on large datasets}. The the overall size of the proxy set is not an important factor for trasferability. CIFAR10 proves to be a valuable proxy set for both ImageNet1K and ImageNet22K. Best viewed in color.}
  \label{fig:transfer_exp} \vspace{-3mm}
\end{figure*}

\vspace{1mm}
\noindent\textbf{Metrics.} Following \cite{Yang2020NAS}, we compute both Top-1 classification accuracy of augmentation runs on target datasets as a metric of performance for each method, and relative improvement (RI) over a random sampling baseline, which is computed as $RI = 100 \times \frac{Acc_m - Acc_r}{Acc_r}$, 
where $Acc_m$ and $Acc_r$ represent Top-1 accuracy of the search method and random sampling, respectively.
RI provides a measure of the quality of each search strategy alone, since both searched and randomly sampled architectures share the same search space and training protocol. A good general-purpose NAS method is expected to yield $RI > 0$ consistently over different searches and across different sub tasks. Note that this comparison is not against random search, but rather against random sampling, i.e., the average architecture of the search space. In our experiments, we compute $Acc_r$ as the average of augmentation runs over $N$ randomly sampled architectures.

%%%%%%%%% Experiments %%%%%%%%%
\section{Results \& Analysis}
\label{sec:experiments}
In this section, we provide detailed analysis on the architecture transfer performance of various NAS methods under different proxy sets, comparison with the random sampling baseline and effect of hyperparameters on the performance of different NAS methods.

\vspace{1mm}
\noindent\textbf{Transferring Architectures.}  \label{ssec:ex_trasfer}
Despite recent efforts to significantly improve the speed of search algorithms, 
performing direct search on large scale target datasets remains prohibitive, unless extremely powerful resources are utilized. 
For example for NSGANet, the search times on CIFAR10 and CIFAR100 with single-GPU is $96$ GPU-hours on average, while single-GPU direct search on ImageNet1K Proxy 1 would take almost two months (which we estimated based on the progress of a four days long run on CIFAR), over one year and half on the full ImageNet1K and approximately $19$ years on ImageNet22K. Even the fastest search method we analyzed, ENAS ($0.375$ days for CIFAR10), would require over one year on ImageNet22K. 
Therefore the need for effective, small-scale proxy sets that could provide a ranking of searched architectures which remains consistent when transferring to the target large scale datasets becomes crucial. But, how to properly select a proxy-set?
From Figure \ref{fig:transfer_exp}, we observe that the size of the proxy set does not influence the effectiveness of an architecture searched on it and then transferred to a vastly larger set. In fact, even architectures searched on the very small FLOWERS102 and MIT67 sets, with less than 10K images, yield actually better results than searches on CIFAR100 for all search strategies. Only DARTS performs better with search on CIFAR10 than the smaller datasets. It is interesting to see that for most search methods transferring to ImageNet1K from CIFAR10 is more effective than transferring from CIFAR100. The overall number of images in CIFAR10 and CIFAR100 is the same, but the number of examples per class is $5,000$ for the former and only $500$ for the latter (half of the ImageNet1K distribution). When comparing the two CIFAR datasets, ENAS seems to privilege a smaller number of examples per class, while DARTS shows an opposite trend.

We also notice how the domain of the proxy set does not seem to influence architecture performance on the target dataset. Intuitively, one would think that a proxy set built from a subset of the target dataset (ImageNet1K Proxy 1 in the Figure, on the right) will yield better results than a proxy set coming from a different dataset (CIFAR10). That appears not to be the case for the target dataset ImageNet22K, as shown in Figure \ref{fig:transfer_exp} on the right. While both CIFAR10 and ImageNet contain images of natural scenes, there is still a significant difference in terms of image subjects and even resolution. Nonetheless, CIFAR10 proves to be a valuable proxy set for large-scale datasets like ImageNet22K.

\begin{figure*}[t]
    \centering
    \captionsetup[subfigure]{labelformat=empty}
    \begin{subfigure}{0.25\linewidth}
        \includegraphics[width=\linewidth]{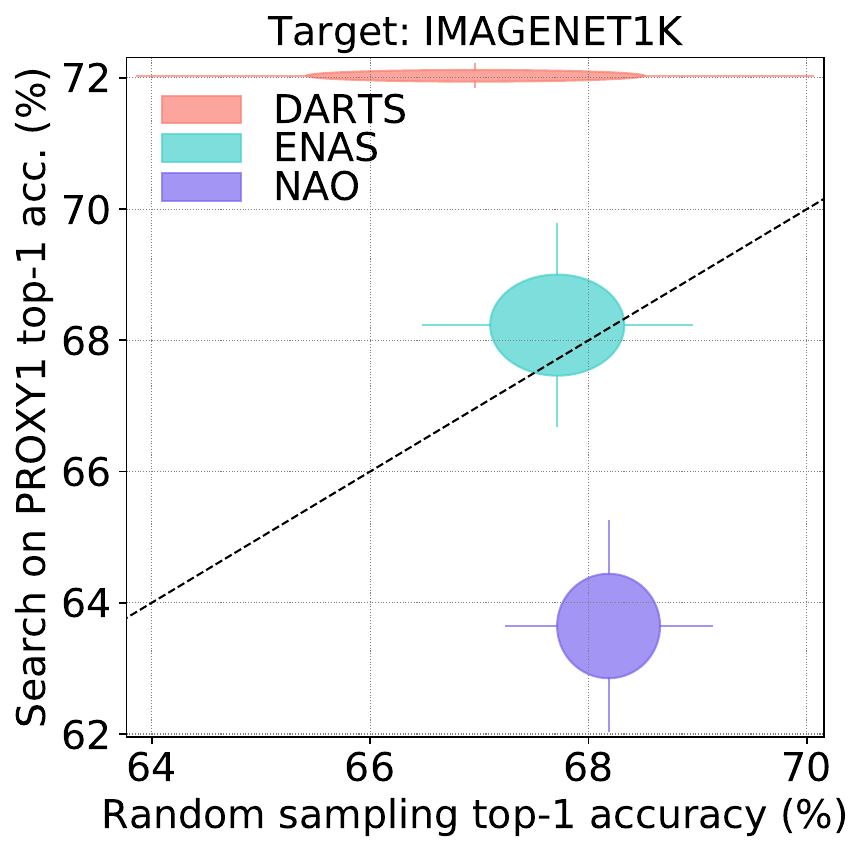}
    \end{subfigure}%
    \begin{subfigure}{0.25\linewidth}
    \includegraphics[width=\linewidth]{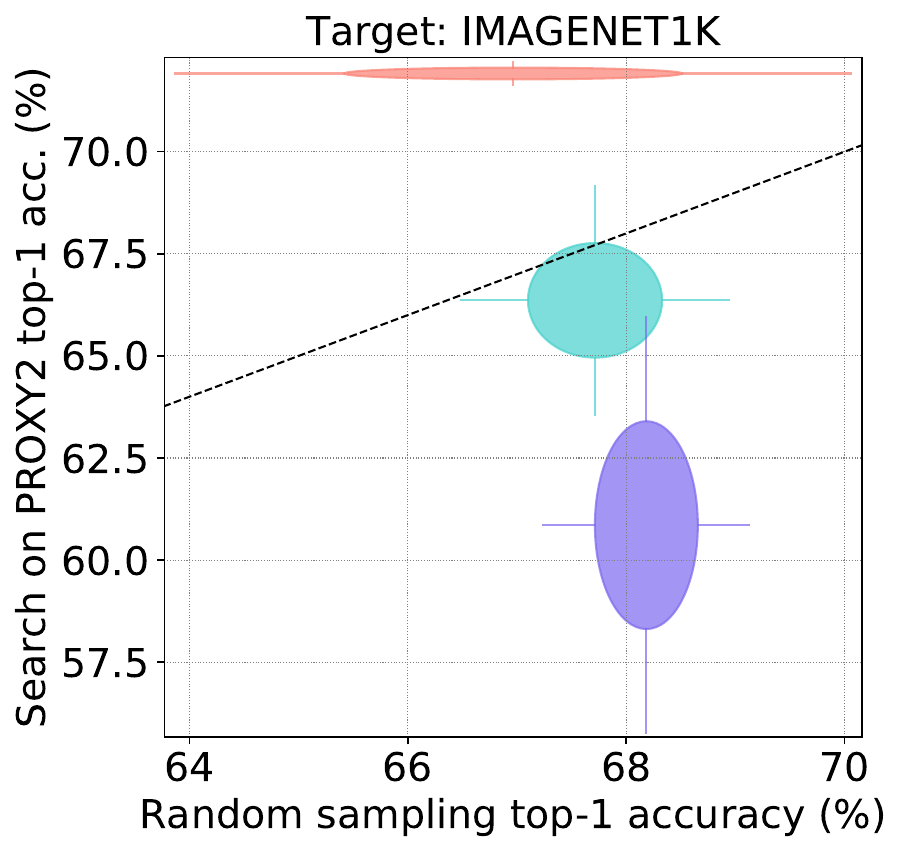}
    \end{subfigure}%
    \begin{subfigure}{0.25\linewidth}
    \includegraphics[width=\linewidth]{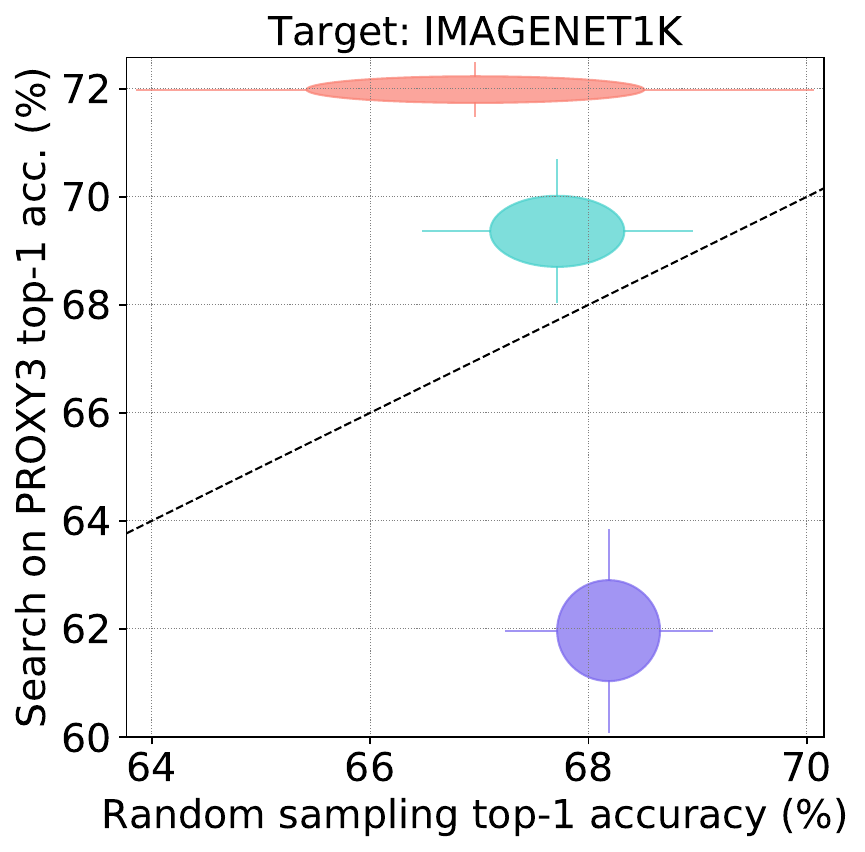}
    \end{subfigure}%
    \begin{subfigure}{0.25\linewidth}
    \includegraphics[width=\linewidth]{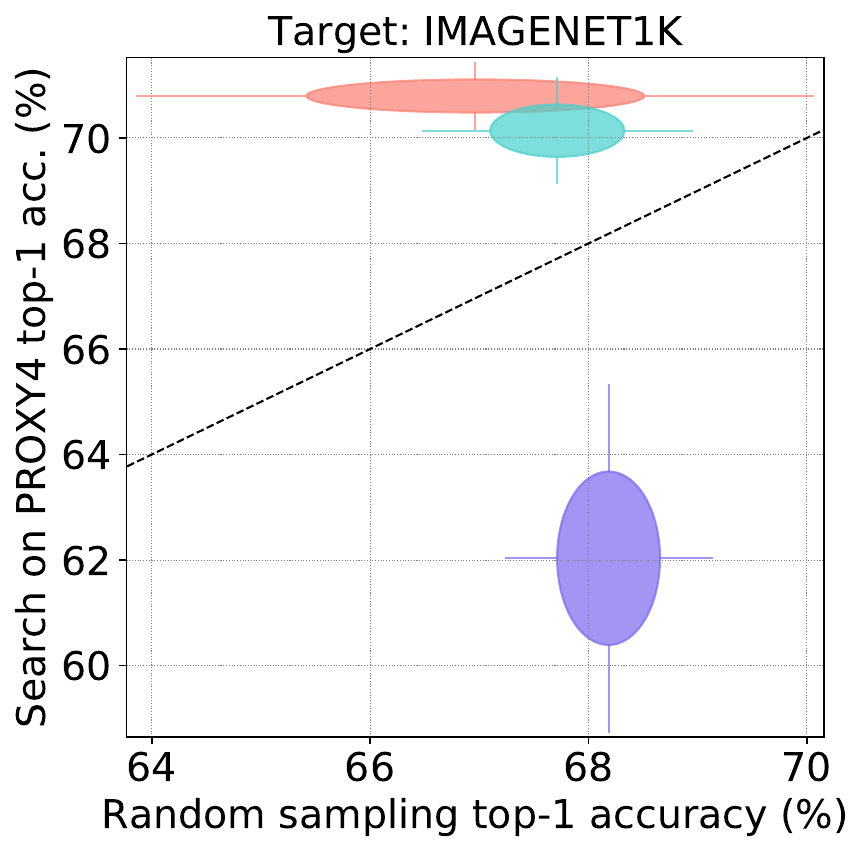}
    \end{subfigure}%

    \begin{subfigure}{0.25\linewidth}
        \includegraphics[width=\linewidth]{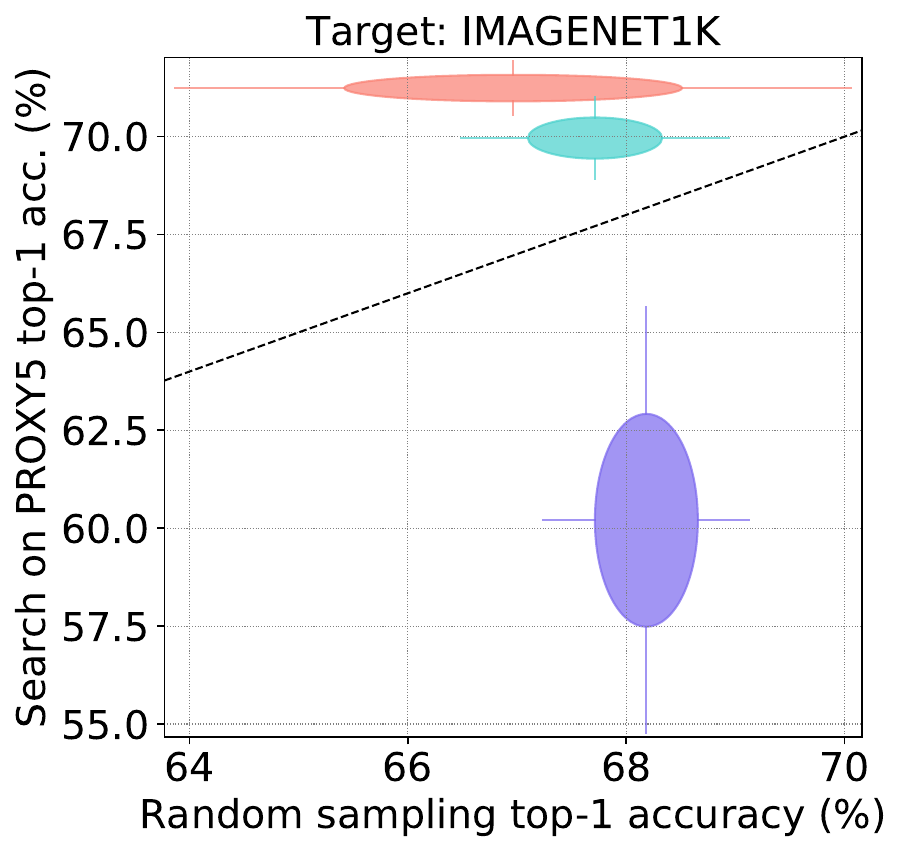}
    \end{subfigure}%
    \begin{subfigure}{0.25\linewidth}
    \includegraphics[width=\linewidth]{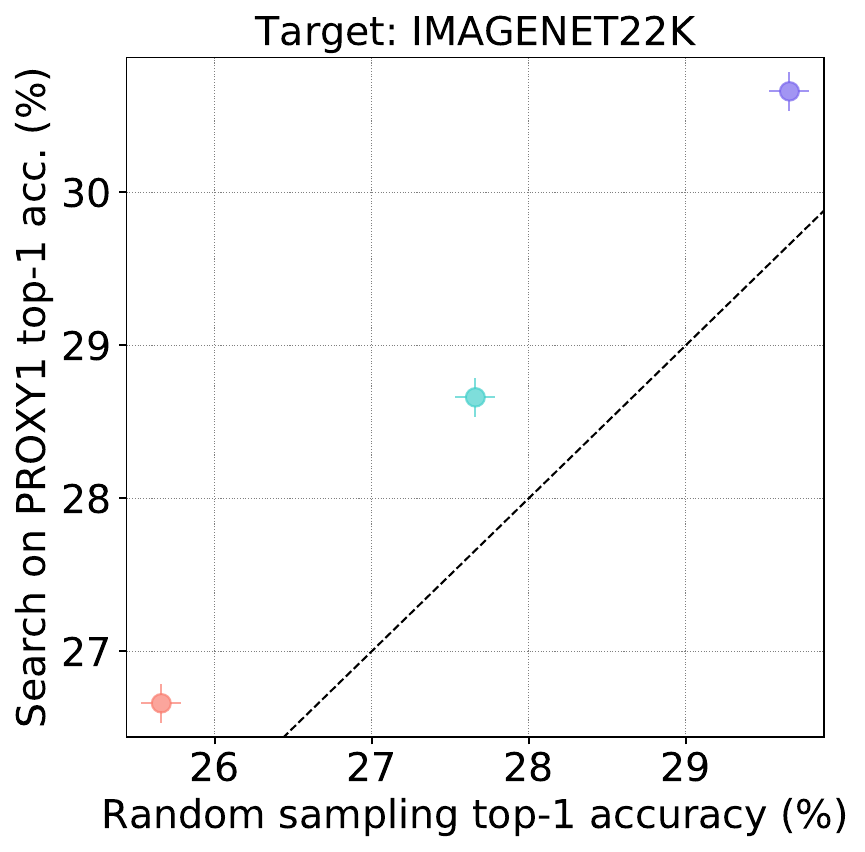}
    \end{subfigure}%
    \begin{subfigure}{0.25\linewidth}
    \includegraphics[width=\linewidth]{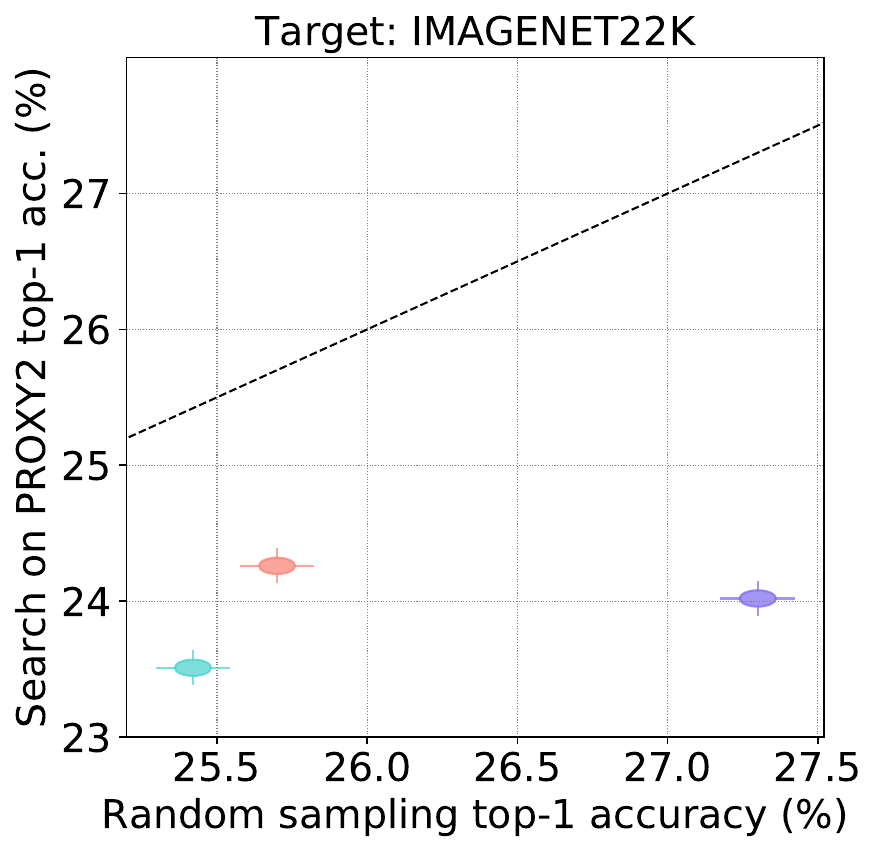}
    \end{subfigure}%
    \begin{subfigure}{0.25\linewidth}
    \includegraphics[width=\linewidth]{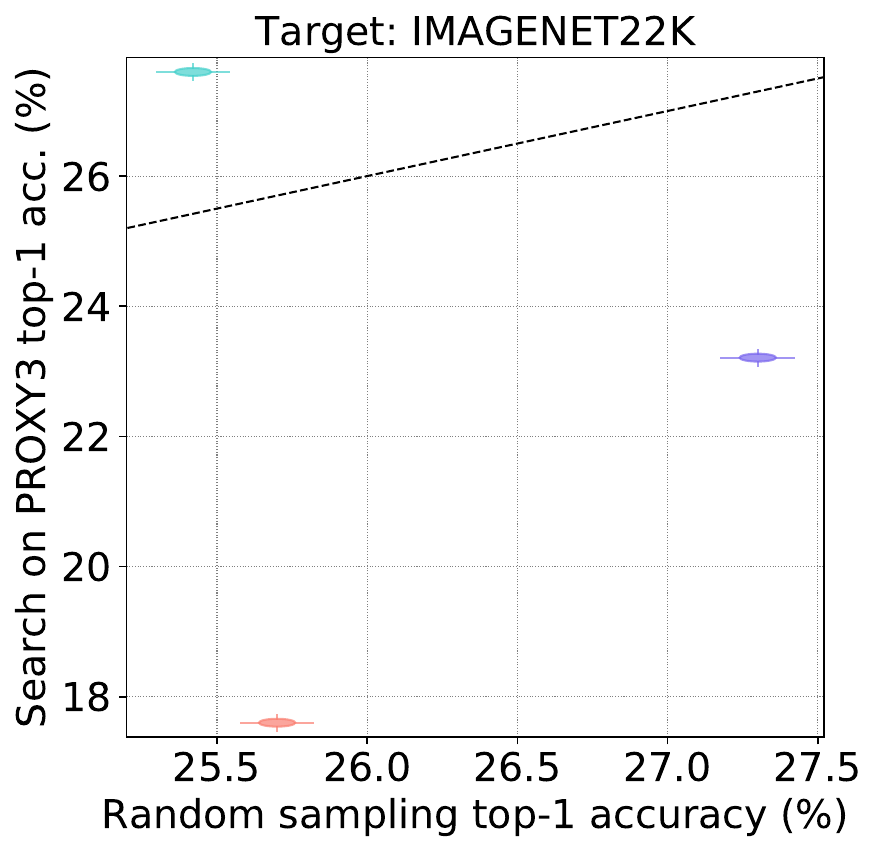}
    \end{subfigure}%
    %\vspace{-1mm}
    \caption{\textbf{Direct search using proxy sets}. Comparison of different NAS methods using sampled proxy sets from same target dataset. Methods lying in diagonal perform the same as randomly sampled architecture, while methods above the diagonal outperform it. We use total five proxy sets for ImageNet1K and three proxy sets for ImageNet22K. Best viewed in color.}
    \label{fig:direct_exp}
    \vspace{-10pt}
\end{figure*}

\vspace{1mm}
\noindent\textbf{Direct Search using Proxy Sets.}
In order to determine the benefit of employing proxy sets directly sampled from the target datasets for architecture search, we compared the performance of the searched architectures versus randomly sampled ones for each of the target datasets. For each method, including random sampling, search was conducted five times, and the resulting mean and standard deviations of the five runs are reported in Figure \ref{fig:direct_exp}.
We observe that for medium scale, uniformly distributed datasets such as ImageNet1K, the rankings of search strategies remains unaffected by proxy set design as DARTS is the best performing method, followed by ENAS and NAO. The number of classes and examples in the proxy set does not bring noticeable improvements, as long as a minimum is guaranteed, as shown by the comparable results using Proxy 1, 2 and 3.
Overall, random sampling of a reduced number classes when building the proxy set seems to provide better performance than keeping all the classes in the target dataset and reducing the number of examples per class (Proxy 5). 

While analyzing the search for very large scale datasets with a skewed distribution (ImageNet22K), the design of proxy set has a large impact not only on the overall improvement, but also on the rankings of different NAS methods. ImageNet22K Proxy 1 results are significantly superior to all other proxies sampled from ImageNet22K for DARTS, ENAS and NAO, and better than random sampling. When searching on ImageNet22K proxies 2 and 3, the random sampling baseline becomes difficult to beat for all methods, and DARTS goes from being the top ranked to the bottom one. This underlines the importance of carefully selecting and designing the proxy set for reliable architecture search. Random sampling of a subset of classes while maintaining the number of images per class proves to be more beneficial, than trying to keep all classes represented in the dataset and eliminating a large portion of examples per class to maintain the search time practically feasible.

% \begin{figure}
%     \centering
%     \includegraphics[width=\textwidth]{fig/direct.pdf}
%     \caption{Direct experiment: Search for architectures on proxy sets of large-scale datasets. Architectures found are trained and tested on the full dataset.}
%     \label{fig:direct_exp}
% \end{figure}

\begin{figure*}[t]
    \centering
    \captionsetup[subfigure]{labelformat=empty}
    \begin{subfigure}{0.31\linewidth}
        \includegraphics[width=\linewidth]{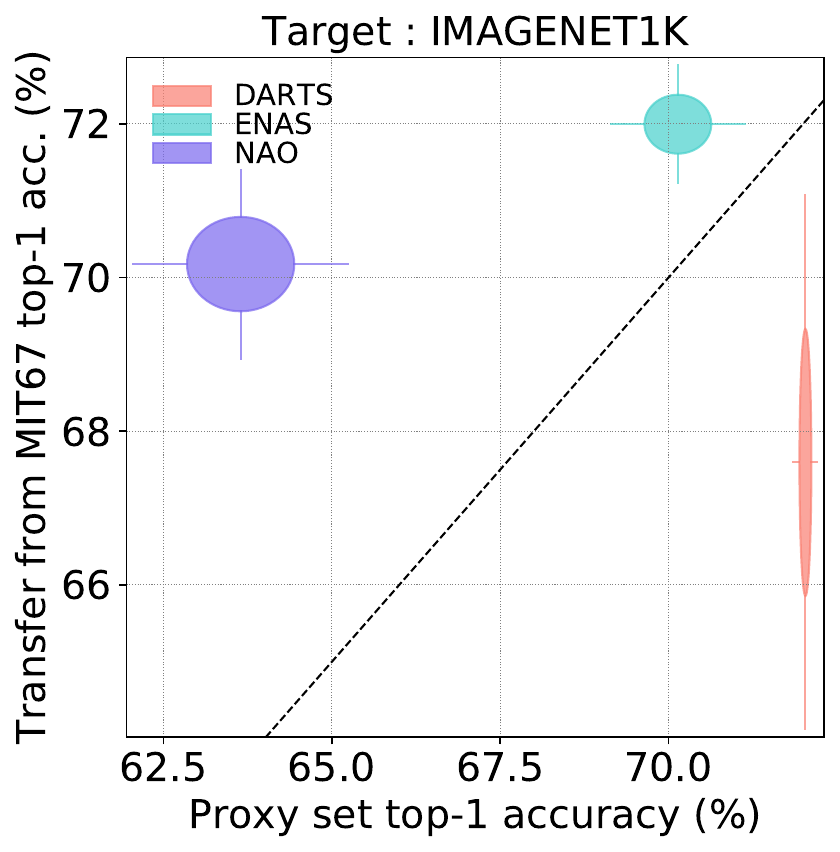}
    \end{subfigure}%
    \begin{subfigure}{0.32\linewidth}
        \includegraphics[width=\linewidth]{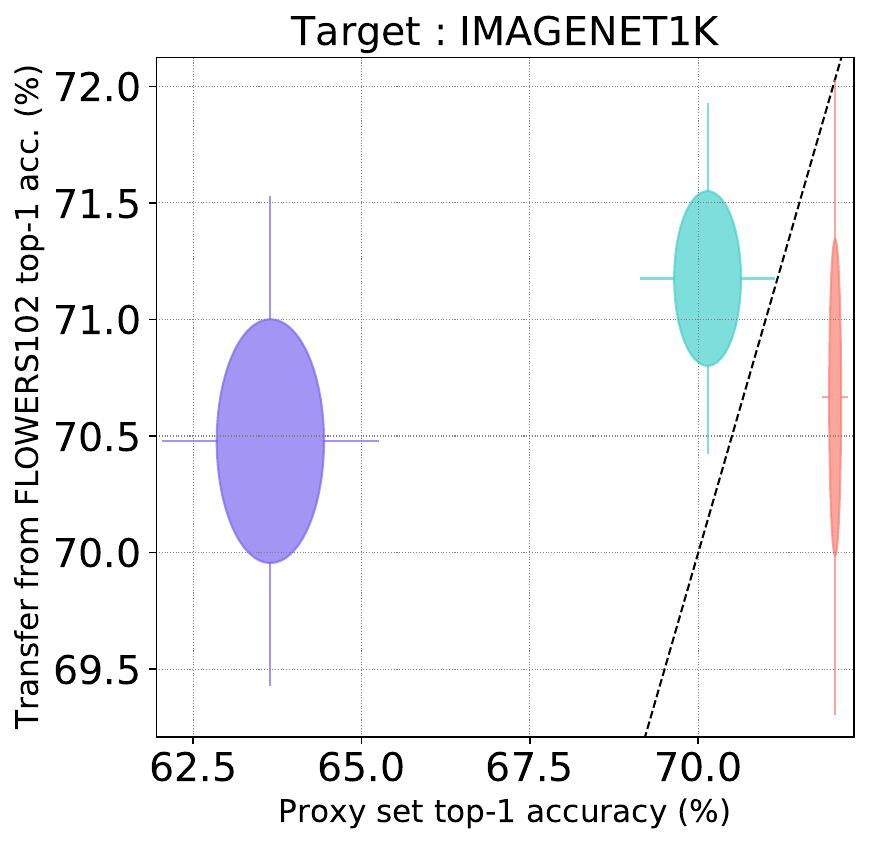}
    \end{subfigure}%
    \begin{subfigure}{0.31\linewidth}
        \includegraphics[width=\linewidth]{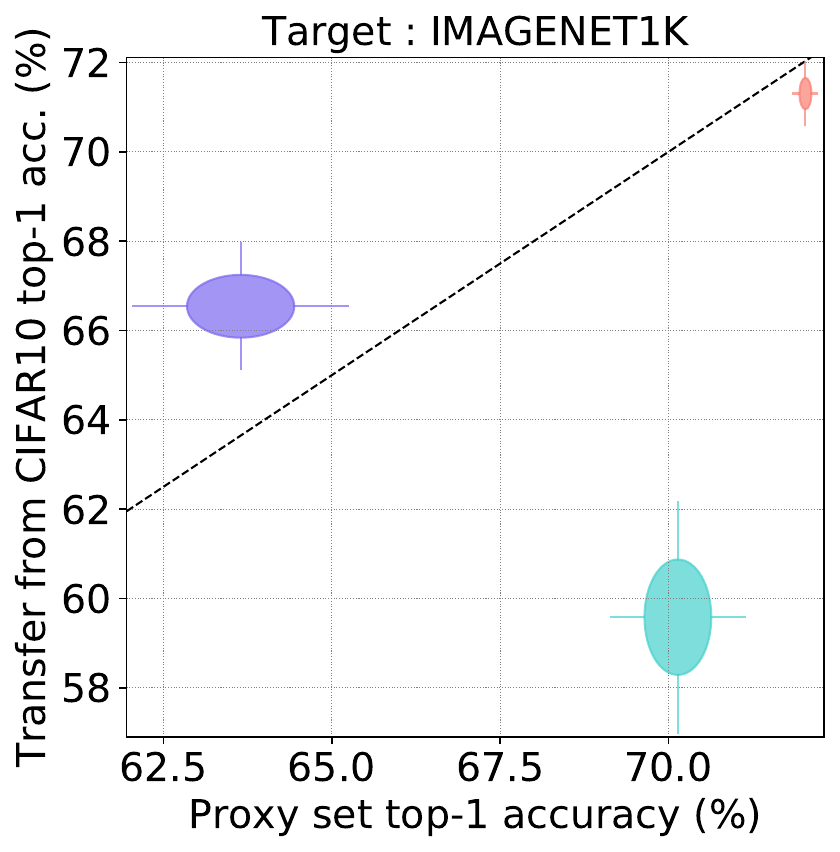}
    \end{subfigure}%
    % 
    
    % \begin{subfigure}{0.3\linewidth}
    % \includegraphics[width=\linewidth]{fig/2_Transfer_vs_search_CIFAR10_to_CIFAR100.pdf}
    % \end{subfigure}%
    \begin{subfigure}{0.31\linewidth}
    \includegraphics[width=\linewidth]{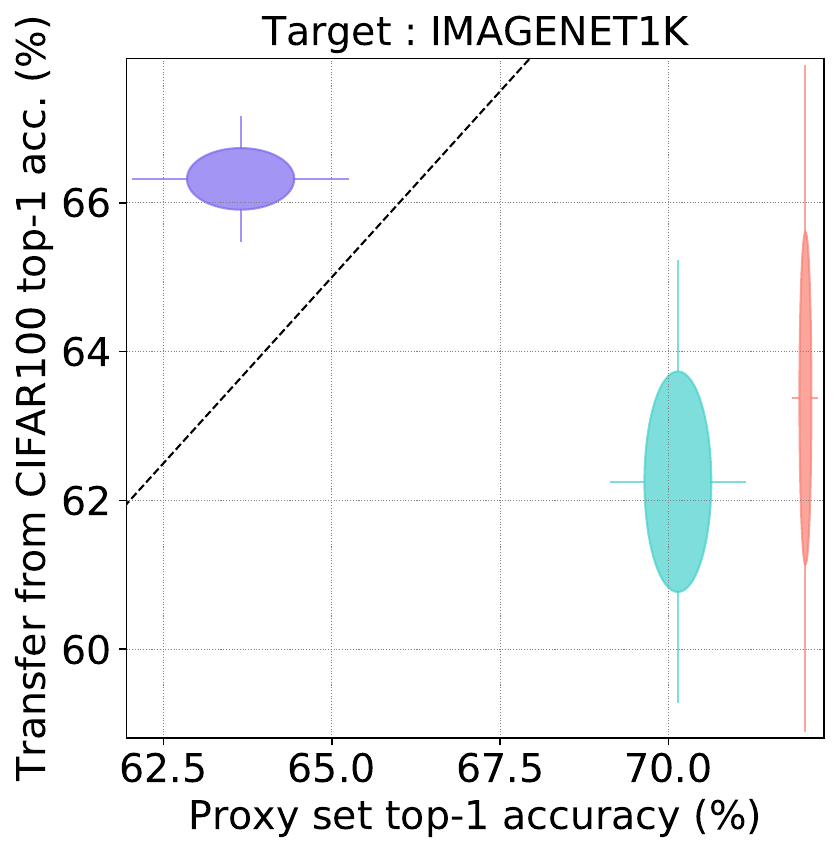}
    \end{subfigure}%
    \begin{subfigure}{0.31\linewidth}
    \includegraphics[width=\linewidth]{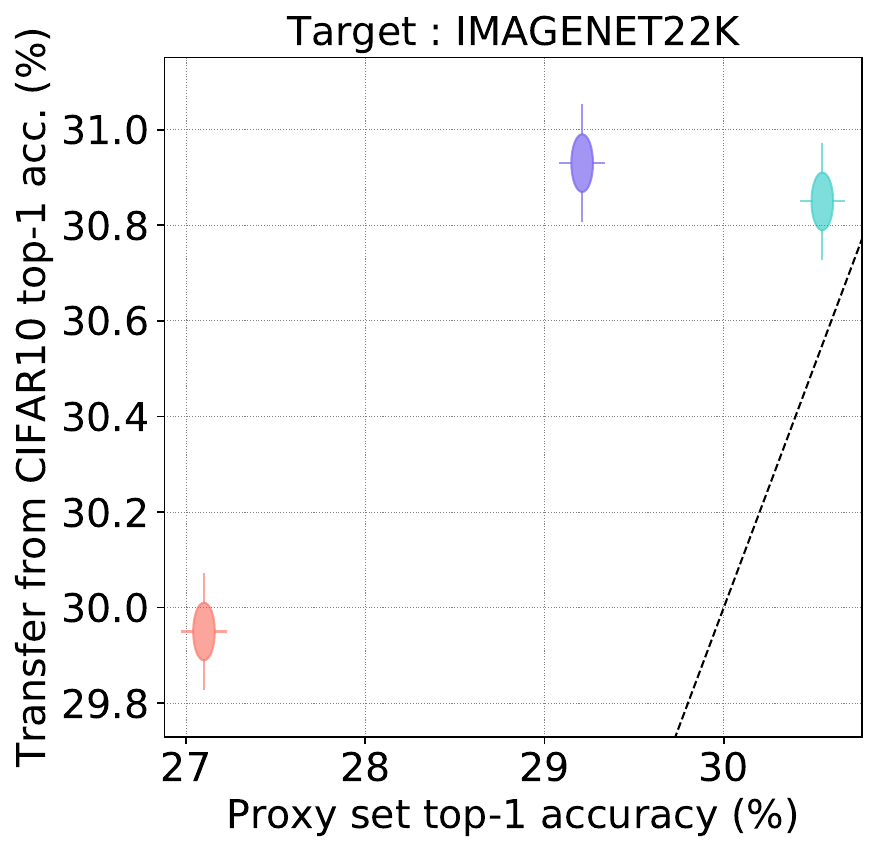}
    \end{subfigure}%
    \begin{subfigure}{0.31\linewidth}
    \includegraphics[width=\linewidth]{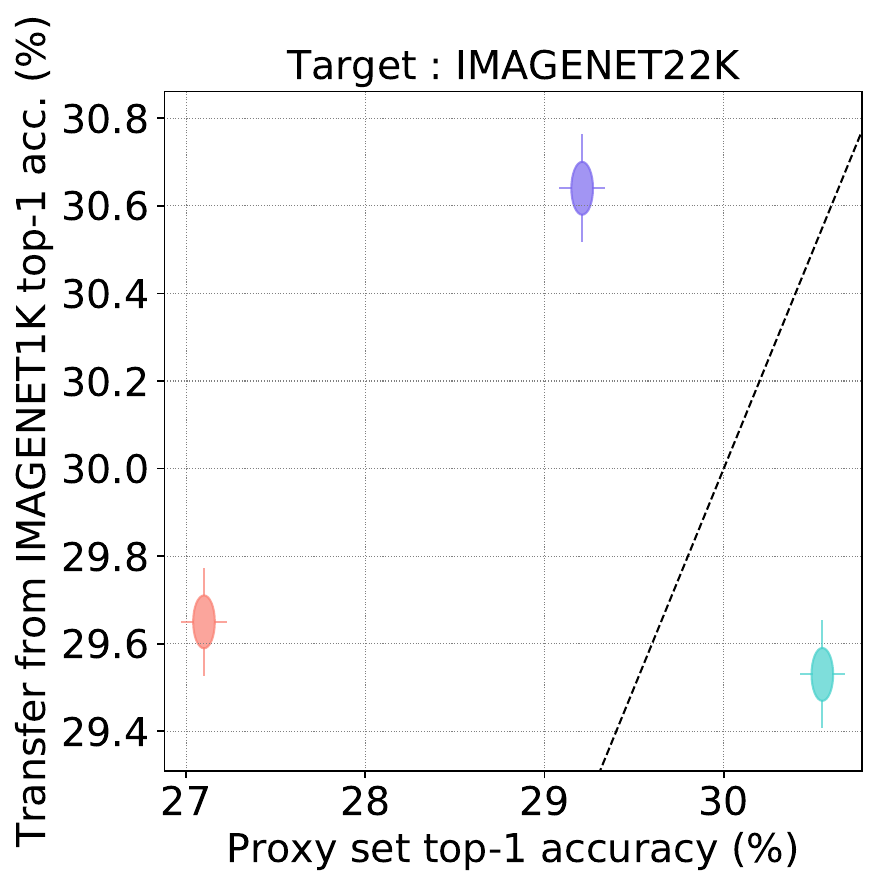}
    \end{subfigure}%
    \caption{\textbf{Architecture transfer vs Proxy-based direct search}. Comparison of transfer performance with the best proxy-based direct search on ImageNet1K and ImageNet22K datasets. Methods lying in the diagonal indicate that transfer performance is similar to the direct proxy-based search, while methods above the diagonal outperform it. Best viewed in color.}
    \label{fig:transfer_vs_direct_exp}
    \vspace{-10pt}
\end{figure*}

\vspace{1mm}
\noindent\textbf{Architecture Transfer vs Proxy-based Direct Search.} 
From the results reported in Figure \ref{fig:transfer_vs_direct_exp}, we can see that on average, transfer performance of architectures searched using completely different small datasets (e.g., MIT67, CIFAR10) can perform similarly or even better than architectures searched directly on proxy target datasets. However, design of proxy sets has considerable impact on rankings of different NAS methods. From the Figure, we can see that for NAO, using a small, unrelated small dataset as proxy provides consistently better results than the best possible proxy sampled from the target dataset, both for ImageNet1K and ImageNet22K. It is surprising to observe how well CIFAR10 works as proxy for ImageNet22k across all methods. Using CIFAR10 produces better results not only than proxy sets from ImageNet1K, but also better than proxy sets directly sampled from ImageNet22K. One would assume that using a subset of the target dataset for search would be beneficial, especially when the distribution across classes is significantly skewed as it is for ImageNet22K. The results of our experiments suggest that a small proxy set, different from the target dataset, albeit in the same general field (natural images classification) can lead the search process to find valuable architectures for datasets even at the scale of ImageNet22K: $14$ million images. For reference, the state-of-the-art published result on ImageNet22K is $36.9$ Top-1 accuracy using a Wide Residual Network  WRN-50-4-2 \cite{intel_blog_2017}, project Adam's network \cite{adamOSDI2014} achieved $29.8\%$, whereas the architecture searched with NAO using CIFAR10 as a proxy yields $30.91$ Top-1 accuracy.

\begin{figure} % 
 \includegraphics[trim={25mm 10mm 25mm 0mm}, width=\linewidth]{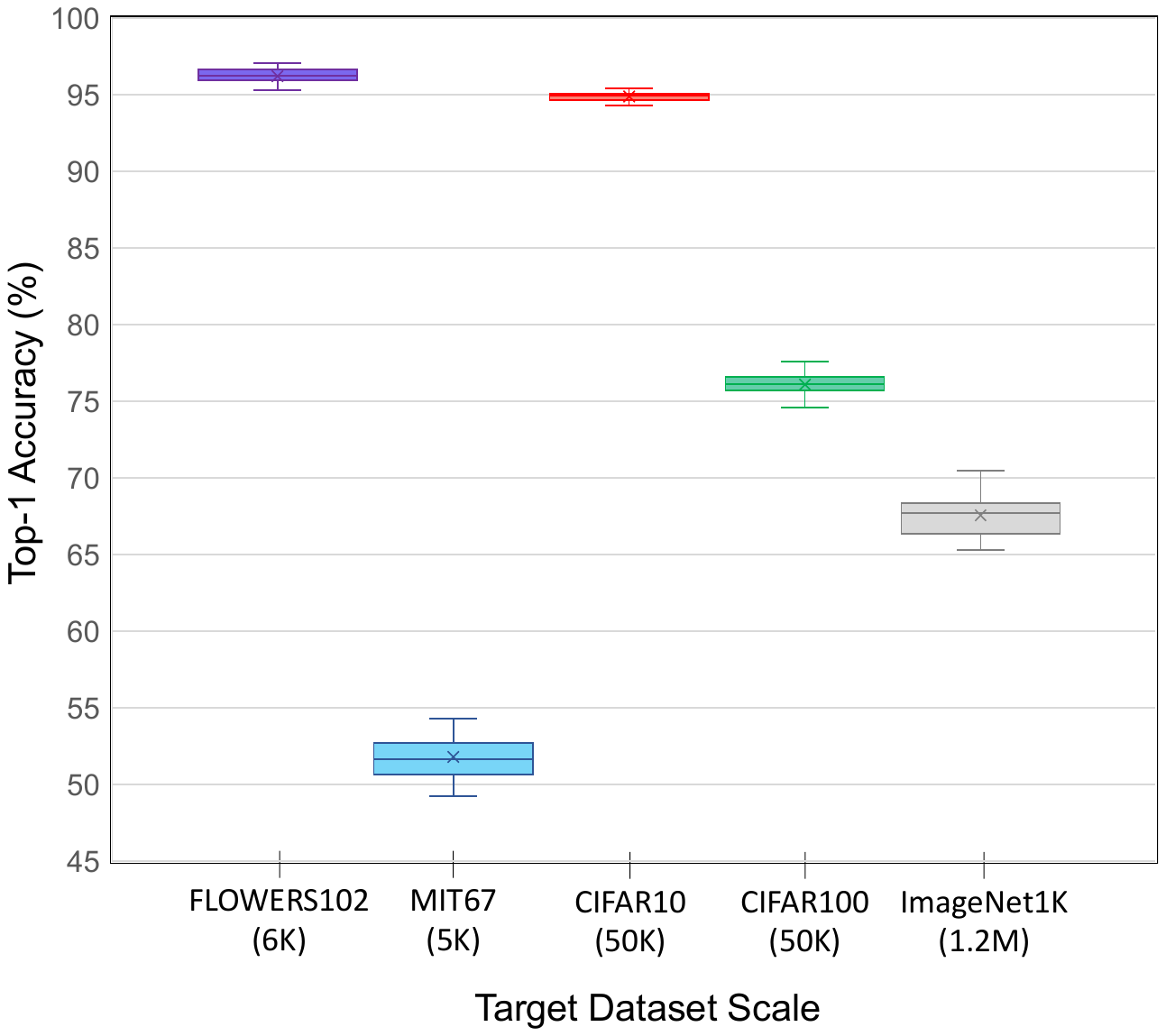}
  \captionof{figure}{\textbf{Random Sampling}. Standard deviation of top-1 accuracy over 30 runs increases with scale and diversity of datasets. Best viewed in color.}
  \label{fig:random_std}
\end{figure}

\begin{table*}[ht]
\centering
\caption{\textbf{Relative improvement metric RI for various transfer experiments}. S and T indicate the source set and target set respectively. 
Given its much longer search time, we did not perform NSGANet search on ImageNet1K proxies.} %\vspace{0.5mm}
\resizebox{\textwidth}{!}{%
\begin{tabular}{c|c|c|c|c|c|c|c}
\toprule
 &
  \begin{tabular}[c]{@{}c@{}}S: CIFAR10 \\ T: CIFAR100\end{tabular} &
  \begin{tabular}[c]{@{}c@{}}S: FLOWERS102 \\ T: ImageNet1K\end{tabular} &
    \begin{tabular}[c]{@{}c@{}}S: MIT67 \\ T: ImageNet1K\end{tabular} &
  \begin{tabular}[c]{@{}c@{}}S: CIFAR10 \\ T: ImageNet1K\end{tabular} &
\begin{tabular}[c]{@{}c@{}}S: CIFAR100 \\ T: ImageNet1K\end{tabular} &
  \begin{tabular}[c]{@{}c@{}}S: CIFAR10 \\ T: ImageNet22K\end{tabular} &
  \begin{tabular}[c]{@{}c@{}}S: ImageNet1K \\ T: ImageNet22K\end{tabular} \\ \hline
NSGANet & 1.38 & -9.16 & -7.02 & 3.39   & 0.92    & -0.72   & -    \\ %\hline
ENAS    & 1.42 & 6.12 & 4.84 & -16.26  & -11.75 & 6.16  & 1.62 \\ %\hline
DARTS   & 1.01 & 6.57 & 0.89 & 5.93   & -18.71  & 3.06 & 2.03 \\ %\hline
NAO     & -0.89 & 2.56 & 1.90 & 0.99   & -2.68 & 6.44  & 5.44    \\ 
\bottomrule
\end{tabular}%
}
\label{tab:relative_improve_rand}
%\vspace{-10pt}
\end{table*}

% \begin{figure}
%     \centering
%     \includegraphics[width=\textwidth]{fig/transfer_vs_direct.pdf}
%     \caption{Comparing performance on transferring architectures searched on smaller datasets versus searching on proxy sets of large-scale datasets.}
%     \label{fig:transfer_vs_direct_exp}
% \end{figure}

\vspace{1mm}
\noindent\textbf{Comparison with Random Sampling.} 
We compare with randomly sampled architectures within the same search space to verify the effectiveness of each method for every augment runs. 
We want to emphasize that this comparison is not against random search, but rather against random sampling, i.e., the average architecture of the search space. We sample $5$ architectures randomly from the search space and compare with the same number of architectures searched by each method.
From the results in Figure \ref{fig:direct_exp} and Table \ref{tab:relative_improve_rand} the random sampling strategy proves to be a very strong baseline, confirming that the effect of a search strategy is less influential for final performance of a given architecture compared to accurately designing the search space. This effect becomes particularly evident when trying to transfer from a small proxy set to a larger one, especially when the number of examples per class varies significantly between proxy and target sets. This is the case for the transfer experiment between CIFAR100 and ImageNet1K, where the architectures learned by all search methods on CIFAR100, with the exception of NSGANet, perform significantly worse than the direct application of randomly sampled ones. Interestingly, CIFAR10 seems instead like a good proxy for transfer to all other target datasets, including ImageNet22K.
%For CIFAR10, CIFAR100 and ImageNet1K we use as baseline for BASE augment settings the average performance of 30 randomly sampled architectures, and DAC augments of 5 randomly sampled ones. 
%For ImageNet22K, given the time consuming nature of running each augment, we randomly sampled 5 architectures for BASE and 5 for DAC. 
To further analyze the effect of number of random sampled architectures, we sample $25$ more randomly sampled architectures (total $30$) on FLOWERS102, MIT67, CIFAR10, CIFAR100 and ImageNet1K.
The influence of search space design and the strength of the random sampling baseline becomes less important as the scale and complexity of the target dataset increases. The standard deviation around the average performance of $30$ randomly sampled architectures expands as the scale of the target dataset increases and as the average accuracy decreases (see Figure \ref{fig:random_std}).  This trend signifies a larger opportunity for impact of the search strategy within the space for datasets that present a hard classification task, like MIT67, and/or large-scale datasets like ImageNet1K.
%We explained in Section \ref{ssec:ex_trasfer} how 
As direct search on large scale datasets is basically intractable in practice, it also shows the importance of finding good proxy sets where search is feasible and performance gains transfer to the target dataset.

\vspace{1mm}
\noindent\textbf{Effect of Hyperparameters.}
We conduct extensive ablation studies over the augmentation training hyperparameters in order to precisely determine the merits of search methods and choice of proxy set versus training protocols. Namely, we investigated augment results by varying seed (Table \ref{tab:aug_seed}) and number of cells (Table \ref{tab:aug_cell}). %More ablation studies are also included in the supplementary material. % and DAC techniques parameters (auxiliary towers weight, cut-out length and drop-path probability, Table \ref{tab:train_protocol}). 
In general we observe that that the parameters of the training protocol have a larger impact on small datasets, but it fails to provide consistent improvements on large datasets, whereas the choice of an appropriate proxy set and search strategy are more relevant.
From Table \ref{tab:aug_seed}, it appears that ENAS is particularly sensitive to choice of augmentation seed, especially when transferring from CIFAR10 to ImageNet1K datset.

\begin{table}[!t]
\centering
\caption{\textbf{Ablation studies on augmentation seed}. Results show performance on two different augmentation seeds. The default value of augmentation seed is set to 2.} 
%\vspace{0.5mm}
\resizebox{0.48\textwidth}{!}{%
\begin{tabular}{c|c|c|c|c} 
\toprule
\multirow{2}{*}{NAS Method} & \multicolumn{2}{c|}{CIFAR10 -- ImageNet1K Transfer}  & \multicolumn{2}{c}{ImageNet1K Proxy 1 Direct}  \\ 
\cmidrule{2-5} & Aug seed = 2   & Aug seed = 3   & Aug seed = 2   & Aug seed = 3   \\ \hline
                  %\cmidrule{2-5} 
NSGANet & 69.86 & 69.27 &    -    &   -        \\ %\hline
ENAS    & 56.58 & 72.69   & 69.40 & 64.7   \\ %\hline
DARTS   & 71.58 & 72.01  & 71.89 & 71.68  \\ %\hline
NAO     & 68.24 & 64.52 & 64.98 & 61.85  \\  %\hline
Random Sampling     & - & - & 70.45 & 71.35 \\ 
\bottomrule
\end{tabular}%
} 
\label{tab:aug_seed}
%\vspace{-10pt}
\end{table}

\begin{table}[!t]
\centering
\caption{\textbf{Effect of number of cells on ImageNet1K}. Results show performance with three different number of cells on ImageNet1K dataset. Increase in number of cells does lead to increase in performance.}
% \caption{\textbf{Effect of number of cells}. Results show performance with three different number of cells on ImageNet1K dataset. Increase in number of cells does lead to increase in performance.}
\resizebox{0.48\textwidth}{!}{%
\begin{tabular}{c|c|c|c|c}
\toprule
\multirow{2}{*}{Experiment}         & \multirow{2}{*}{NAS Method} & \multicolumn{3}{c}{Number of Cells}                                        \\ \cline{3-5} 
                                    &      & 20 & 40 & 60 \\  \hline
\multirow{4}{*}{\begin{tabular}[c]{@{}c@{}}CIFAR10 -- Imagenet1K\\ Transfer\end{tabular}} & NSGANet & 68.38  & 71.62  & 62.16  \\ %\cline{2-8} 
                                    & ENAS                        & 56.58          & 72.5          & 72.55            \\ %\cline{2-8} 
                                    & DARTS                       & 71.58           & 63.75           & 53.98            \\ %\cline{2-8} 
                                    & NAO                         & 68.24           & 63.87        & 58.54               \\ \hline
\multirow{3}{*}{ImageNet1K Proxy 1} & ENAS                        & 69.4             & 69.61          & 60.31        \\ %\cline{2-8} 
                                    & DARTS                       & 71.89           & 70.04            & 60.6             \\ %\cline{2-8} 
                                    & NAO                         & 64.98           & 67.32           & 52.45           \\ \hline %\hline
ImageNet1K &
  \begin{tabular}[c]{@{}c@{}}Random \\ Sampling\end{tabular} &
  70.45 &

  65.52 &

  52.82  \\ 
  \bottomrule
\end{tabular}%
}
\label{tab:aug_cell}
%\vspace{-10pt}
\end{table}

\section{Conclusions and Best Practices}
\label{sec:discussion}
We have presented the first extensive study on the design and transfer value of proxy sets for NAS at large scale across different search methods. We compared four standard search strategies on proxy datasets ranging from small (5K images) to large (384K images) and their transferability to very large scale target sets including for the first time ImageNet22K (14M images).
The results of our experiments and ablation studies suggest the following set of best practices when choosing proxy sets.
(i)  The the overall size of the proxy set is not an important factor for trasferability. Reliably good search for large scale datasets can be performed on proxy sets even smaller than two order of magnitude with respect to the target ones.
(ii) The domain of the proxy set does not seem to influence architecture performance on the target dataset. Transfer performance of architectures searched using different small datasets (e.g., MIT67, CIFAR10) can perform similarly or even better than architectures searched directly on proxies of target datasets (ImageNet1K and ImageNet22K).
(iii) For proxy sets directly sampled from the target set, the random sampling of a subset of classes maintaining the same number of images per class is more beneficial than trying to keep all the classes from the target dataset and the reducing the number of examples per class.
(iv) As the scale of target dataset increases, the choice of proxy set and search strategy matters more on the final augmentation performance on the target dataset than the training protocol and hyper-parameters setting. NAS methods showing similar performance on a source dataset (e.g., CIFAR10), produce largely different transfer performances to a large dataset (e.g. ImagenNet22K).
(v) Random sampling remains a strong baseline to surpass, but the choice of the appropriate combination of proxy set and search strategy can provide significant improvement over it. 
In future, we plan to further study the transferablity from proxy sets of significantly different domains. We also believe that including the appropriate selection of proxy set in combination with a search strategy within a unified NAS framework can lead to not only efficient but also more effective NAS at large scale.

\vspace{1mm}
\noindent\textbf{Acknowledgements.} This research used resources of the Oak Ridge Leadership Computing Facility, which is a DOE Office of Science User Facility supported under Contract DE-AC05-00OR22725. It also used resources of
the IBM T.J. Watson Research Center Scaling Cluster (WSC).

\vspace{1mm}
\noindent\textbf{Disclaimer.} ImageNet was used only for research purposes to allow benchmarking against prior results. The trained models in this work are not used for commercial purposes.

\small \bibliography{egbib}

\begin{thebibliography}{50}
\providecommand{\natexlab}[1]{#1}
\providecommand{\url}[1]{\texttt{#1}}
\providecommand{\urlprefix}{URL }
\expandafter\ifx\csname urlstyle\endcsname\relax
  \providecommand{\doi}[1]{doi:\discretionary{}{}{}#1}\else
  \providecommand{\doi}{doi:\discretionary{}{}{}\begingroup
  \urlstyle{rm}\Url}\fi

\bibitem[{Baker et~al.(2018)Baker, Gupta, Raskar, and
  Naik}]{baker2018accelerating}
Baker, B.; Gupta, O.; Raskar, R.; and Naik, N. 2018.
\newblock Accelerating Neural Architecture Search using Performance Prediction.
\newblock In \emph{ICLR Workshops}.

\bibitem[{Bender et~al.(2018)Bender, Kindermans, Zoph, Vasudevan, and
  Le}]{pmlr-v80-bender18a}
Bender, G.; Kindermans, P.-J.; Zoph, B.; Vasudevan, V.; and Le, Q. 2018.
\newblock Understanding and Simplifying One-Shot Architecture Search.
\newblock In \emph{ICML}.

\bibitem[{Borsos, Khorlin, and Gesmundo(2019)}]{Borsos2019TransferNK}
Borsos, Z.; Khorlin, A.; and Gesmundo, A. 2019.
\newblock Transfer NAS: Knowledge Transfer between Search Spaces with
  Transformer Agents.
\newblock \emph{arXiv preprint 1906.08102} .

\bibitem[{Brock et~al.(2018)Brock, Lim, Ritchie, and Weston}]{brock2018smash}
Brock, A.; Lim, T.; Ritchie, J.; and Weston, N. 2018.
\newblock {SMASH}: One-Shot Model Architecture Search through HyperNetworks.
\newblock In \emph{ICLR}.

\bibitem[{Cai, Zhu, and Han(2019)}]{cai2019proxylessnas}
Cai, H.; Zhu, L.; and Han, S. 2019.
\newblock Proxyless{NAS}: Direct Neural Architecture Search on Target Task and
  Hardware.
\newblock In \emph{ICLR}.

\bibitem[{Chang et~al.(2019)Chang, zhang, Guo, MENG, XIANG, and
  Pan}]{DATA_neurips2019}
Chang, J.; zhang, x.; Guo, Y.; MENG, G.; XIANG, S.; and Pan, C. 2019.
\newblock DATA: Differentiable ArchiTecture Approximation.
\newblock In \emph{Advances in Neural Information Processing Systems}.

\bibitem[{Chen et~al.(2019)Chen, Yang, Zhang, MENG, Xiao, and
  Sun}]{DetNAS_Neurips2019}
Chen, Y.; Yang, T.; Zhang, X.; MENG, G.; Xiao, X.; and Sun, J. 2019.
\newblock DetNAS: Backbone Search for Object Detection.
\newblock In \emph{Advances in Neural Information Processing Systems}.

\bibitem[{Chilimbi et~al.(2014)Chilimbi, Suzue, Apacible, and
  Kalyanaraman}]{adamOSDI2014}
Chilimbi, T.; Suzue, Y.; Apacible, J.; and Kalyanaraman, K. 2014.
\newblock Project Adam: Building an Efficient and Scalable Deep Learning
  Training System.
\newblock In \emph{11th {USENIX} Symposium on Operating Systems Design and
  Implementation ({OSDI})}, 571--582.

\bibitem[{Cho et~al.(2017)Cho, Finkler, Kumar, Kung, Saxena, and
  Sreedhar}]{cho2017powerai}
Cho, M.; Finkler, U.; Kumar, S.; Kung, D.; Saxena, V.; and Sreedhar, D. 2017.
\newblock PowerAI DDL.
\newblock In \emph{arXiv preprint 1708.02188}.

\bibitem[{Codreanu, Podareanu, and Saletore(2017)}]{intel_blog_2017}
Codreanu, V.; Podareanu, D.; and Saletore, V. 2017.
\newblock
  \url{https://communities.surf.nl/artikel/achieving-deep-learning-training-in-less-than-40-minutes-on-imagenet-1k-best-accuracy-and}.

\bibitem[{Deng et~al.(2009)Deng, Dong, Socher, Li, Li, and
  Fei-Fei}]{imagenet_cvpr09}
Deng, J.; Dong, W.; Socher, R.; Li, L.-J.; Li, K.; and Fei-Fei, L. 2009.
\newblock {ImageNet: A Large-Scale Hierarchical Image Database}.
\newblock In \emph{CVPR}.

\bibitem[{DeVries and Taylor(2017)}]{devries2017cutout}
DeVries, T.; and Taylor, G.~W. 2017.
\newblock Improved Regularization of Convolutional Neural Networks with Cutout.
\newblock \emph{arXiv preprint 1708.04552} .

\bibitem[{Dong and Yang(2020)}]{Dong2020NAS-Bench-201}
Dong, X.; and Yang, Y. 2020.
\newblock NAS-Bench-201: Extending the Scope of Reproducible Neural
  Architecture Search.
\newblock In \emph{ICLR}.

\bibitem[{Elsken, Metzen, and Hutter(2019)}]{survey_JMLR2019}
Elsken, T.; Metzen, J.~H.; and Hutter, F. 2019.
\newblock Neural Architecture Search: A Survey.
\newblock \emph{JMLR} 20(55): 1--21.

\bibitem[{Finkler et~al.(2020)Finkler, Merler, Panda, Jaiswal, Wu,
  Ramakrishnan, Chen, Cho, Kung, Feris et~al.}]{finkler2020large}
Finkler, U.; Merler, M.; Panda, R.; Jaiswal, M.~S.; Wu, H.; Ramakrishnan, K.;
  Chen, C.-F.; Cho, M.; Kung, D.; Feris, R.; et~al. 2020.
\newblock Large Scale Neural Architecture Search with Polyharmonic Splines.
\newblock \emph{arXiv preprint arXiv:2011.10608} .

\bibitem[{{Ghiasi}, {Lin}, and {Le}(2019)}]{NASFPN_CVPR2019}
{Ghiasi}, G.; {Lin}, T.; and {Le}, Q.~V. 2019.
\newblock NAS-FPN: Learning Scalable Feature Pyramid Architecture for Object
  Detection.
\newblock In \emph{CVPR}.

\bibitem[{Krizhevsky(2009)}]{Krizhevsky09learningmultiple}
Krizhevsky, A. 2009.
\newblock Learning multiple layers of features from tiny images.
\newblock Technical report.

\bibitem[{Kyriakides and Margaritis(2020)}]{reduced_training_NCA2020}
Kyriakides, G.; and Margaritis, K. 2020.
\newblock The effect of reduced training in neural architecture search.
\newblock \emph{Neural Computing and Applications} .

\bibitem[{Larsson, Maire, and Shakhnarovich(2017)}]{larsson2017fractalnet}
Larsson, G.; Maire, M.; and Shakhnarovich, G. 2017.
\newblock FractalNet: Ultra-Deep Neural Networks without Residuals.
\newblock In \emph{ICLR}.

\bibitem[{Li and Talwalkar(2019)}]{LiTUAI19}
Li, L.; and Talwalkar, A. 2019.
\newblock Random Search and Reproducibility for Neural Architecture Search.
\newblock In \emph{Conference on Uncertainty in Artificial Intelligence {UAI}}.

\bibitem[{Lian et~al.(2020)Lian, Zheng, Xu, Lu, Lin, Zhao, Huang, and
  Gao}]{Lian2020Towards}
Lian, D.; Zheng, Y.; Xu, Y.; Lu, Y.; Lin, L.; Zhao, P.; Huang, J.; and Gao, S.
  2020.
\newblock Towards Fast Adaptation of Neural Architectures with Meta Learning.
\newblock In \emph{ICLR}.

\bibitem[{Liu et~al.(2019)Liu, Chen, Schroff, Adam, Hua, Yuille, and
  Fei-Fei}]{liu2019autodeeplab}
Liu, C.; Chen, L.-C.; Schroff, F.; Adam, H.; Hua, W.; Yuille, A.; and Fei-Fei,
  L. 2019.
\newblock Auto-DeepLab: Hierarchical Neural Architecture Search for Semantic
  Image Segmentation.
\newblock In \emph{CVPR}.

\bibitem[{Liu et~al.(2018{\natexlab{a}})Liu, Zoph, Neumann, Shlens, Hua, Li,
  Fei-Fei, Yuille, Huang, and Murphy}]{progressiveNAS_ECCV2018}
Liu, C.; Zoph, B.; Neumann, M.; Shlens, J.; Hua, W.; Li, L.-J.; Fei-Fei, L.;
  Yuille, A.; Huang, J.; and Murphy, K. 2018{\natexlab{a}}.
\newblock Progressive Neural Architecture Search.
\newblock In \emph{ECCV}.

\bibitem[{Liu et~al.(2018{\natexlab{b}})Liu, Simonyan, Vinyals, Fernando, and
  Kavukcuoglu}]{liu2018hierarchical}
Liu, H.; Simonyan, K.; Vinyals, O.; Fernando, C.; and Kavukcuoglu, K.
  2018{\natexlab{b}}.
\newblock Hierarchical Representations for Efficient Architecture Search.
\newblock In \emph{ICLR}.

\bibitem[{Liu, Simonyan, and Yang(2018)}]{liu2018darts}
Liu, H.; Simonyan, K.; and Yang, Y. 2018.
\newblock DARTS: Differentiable Architecture Search.
\newblock \emph{arXiv preprint 1806.09055} .

\bibitem[{Lu et~al.(2020)Lu, Sreekumar, Goodman, Banzhaf, Deb, and
  Boddeti}]{lu2020neural}
Lu, Z.; Sreekumar, G.; Goodman, E.; Banzhaf, W.; Deb, K.; and Boddeti, V.~N.
  2020.
\newblock Neural Architecture Transfer.
\newblock \emph{arXiv preprint 2005.05859} .

\bibitem[{Lu et~al.(2019)Lu, Whalen, Boddeti, Dhebar, Deb, Goodman, and
  Banzhaf}]{NSGA_Net_2019}
Lu, Z.; Whalen, I.; Boddeti, V.; Dhebar, Y.; Deb, K.; Goodman, E.; and Banzhaf,
  W. 2019.
\newblock NSGA-Net: Neural Architecture Search Using Multi-Objective Genetic
  Algorithm.
\newblock In \emph{Proceedings of the Genetic and Evolutionary Computation
  Conference}.

\bibitem[{Luo et~al.(2018)Luo, Tian, Qin, Chen, and Liu}]{NAO}
Luo, R.; Tian, F.; Qin, T.; Chen, E.-H.; and Liu, T.-Y. 2018.
\newblock Neural Architecture Optimization.
\newblock In \emph{Advances in neural information processing systems}.

\bibitem[{Nayman et~al.(2019)Nayman, Noy, Ridnik, Friedman, Jin, and
  Zelnik-Manor}]{nayman2019xnas}
Nayman, N.; Noy, A.; Ridnik, T.; Friedman, I.; Jin, R.; and Zelnik-Manor, L.
  2019.
\newblock XNAS: Neural Architecture Search with Expert Advice.
\newblock In \emph{Advances in Neural Information Processing Systems}.

\bibitem[{Nilsback and Zisserman(2008)}]{nilsback2008automated}
Nilsback, M.-E.; and Zisserman, A. 2008.
\newblock Automated flower classification over a large number of classes.
\newblock In \emph{2008 Sixth Indian Conference on Computer Vision, Graphics \&
  Image Processing}.

\bibitem[{Pham et~al.(2018)Pham, Guan, Zoph, Le, and Dean}]{pmlr-v80-pham18a}
Pham, H.; Guan, M.; Zoph, B.; Le, Q.; and Dean, J. 2018.
\newblock Efficient Neural Architecture Search via Parameters Sharing.
\newblock In \emph{ICML}.

\bibitem[{Quattoni and Torralba(2009)}]{quattoni2009recognizing}
Quattoni, A.; and Torralba, A. 2009.
\newblock Recognizing indoor scenes.
\newblock In \emph{CVPR}.

\bibitem[{Real et~al.(2019)Real, Aggarwal, Huang, and
  Le}]{regularized_AAAI2019}
Real, E.; Aggarwal, A.; Huang, Y.; and Le, Q.~V. 2019.
\newblock Regularized Evolution for Image Classifier Architecture Search.
\newblock In \emph{AAAI}.

\bibitem[{Russakovsky et~al.(2015)Russakovsky, Deng, Su, Krause, Satheesh, Ma,
  Huang, Karpathy, Khosla, Bernstein, Berg, and Fei-Fei}]{ILSVRC15}
Russakovsky, O.; Deng, J.; Su, H.; Krause, J.; Satheesh, S.; Ma, S.; Huang, Z.;
  Karpathy, A.; Khosla, A.; Bernstein, M.; Berg, A.~C.; and Fei-Fei, L. 2015.
\newblock ImageNet Large Scale Visual Recognition Challenge.
\newblock \emph{IJCV} 115(3): 211--252.

\bibitem[{{Tan} et~al.(2019){Tan}, {Chen}, {Pang}, {Vasudevan}, {Sandler},
  {Howard}, and {Le}}]{MnasNet_CVPR2019}
{Tan}, M.; {Chen}, B.; {Pang}, R.; {Vasudevan}, V.; {Sandler}, M.; {Howard},
  A.; and {Le}, Q.~V. 2019.
\newblock MnasNet: Platform-Aware Neural Architecture Search for Mobile.
\newblock In \emph{CVPR}.

\bibitem[{Wang et~al.(2019)Wang, Gao, Chen, Wang, Tian, and
  Shen}]{wang2019nasfcos}
Wang, N.; Gao, Y.; Chen, H.; Wang, P.; Tian, Z.; and Shen, C. 2019.
\newblock {NAS-FCOS}: Fast Neural Architecture Search for Object Detection.
\newblock \emph{arXiv preprint 1906.04423} .

\bibitem[{Wistuba(2019)}]{wistuba2019xfernas}
Wistuba, M. 2019.
\newblock XferNAS: Transfer Neural Architecture Search.
\newblock \emph{arXiv preprint 1907.08307} .

\bibitem[{Wu et~al.(2019)Wu, Dai, Zhang, Wang, Sun, Wu, Tian, Vajda, Jia, and
  Keutzer}]{wu2018fbnet}
Wu, B.; Dai, X.; Zhang, P.; Wang, Y.; Sun, F.; Wu, Y.; Tian, Y.; Vajda, P.;
  Jia, Y.; and Keutzer, K. 2019.
\newblock FBNet: Hardware-Aware Efficient ConvNet Design via Differentiable
  Neural Architecture Search.
\newblock In \emph{CVPR}.

\bibitem[{{Xie} et~al.(2019){Xie}, {Kirillov}, {Girshick}, and
  {He}}]{randomly_wired_ICCV2019}
{Xie}, S.; {Kirillov}, A.; {Girshick}, R.; and {He}, K. 2019.
\newblock Exploring Randomly Wired Neural Networks for Image Recognition.
\newblock In \emph{ICCV}.

\bibitem[{Xie et~al.(2019)Xie, Zheng, Liu, and Lin}]{xie2019snas}
Xie, S.; Zheng, H.; Liu, C.; and Lin, L. 2019.
\newblock {SNAS}: stochastic neural architecture search.
\newblock In \emph{ICLR}.

\bibitem[{Yang, Esperança, and Carlucci(2020)}]{Yang2020NAS}
Yang, A.; Esperança, P.~M.; and Carlucci, F.~M. 2020.
\newblock NAS evaluation is frustratingly hard.
\newblock In \emph{ICLR}.

\bibitem[{Ying et~al.(2019)Ying, Klein, Christiansen, Real, Murphy, and
  Hutter}]{pmlr-v97-ying19a}
Ying, C.; Klein, A.; Christiansen, E.; Real, E.; Murphy, K.; and Hutter, F.
  2019.
\newblock {NAS}-Bench-101: Towards Reproducible Neural Architecture Search.
\newblock In \emph{ICML}.

\bibitem[{Yu et~al.(2020)Yu, Sciuto, Jaggi, Musat, and
  Salzmann}]{Yu2020Evaluating}
Yu, K.; Sciuto, C.; Jaggi, M.; Musat, C.; and Salzmann, M. 2020.
\newblock Evaluating The Search Phase of Neural Architecture Search.
\newblock In \emph{ICLR}.

\bibitem[{Zela et~al.(2020)Zela, Elsken, Saikia, Marrakchi, Brox, and
  Hutter}]{Zela2020Understanding}
Zela, A.; Elsken, T.; Saikia, T.; Marrakchi, Y.; Brox, T.; and Hutter, F. 2020.
\newblock Understanding and Robustifying Differentiable Architecture Search.
\newblock In \emph{ICLR}.

\bibitem[{Zela, Siems, and Hutter(2020)}]{Zela2020NAS-Bench-1Shot1}
Zela, A.; Siems, J.; and Hutter, F. 2020.
\newblock NAS-Bench-1Shot1: Benchmarking and Dissecting One-shot Neural
  Architecture Search.
\newblock In \emph{ICLR}.

\bibitem[{Zhang et~al.(2015)Zhang, Hu, Wei, Xie, Kim, Ho, and
  Xing}]{zhang2015poseidon}
Zhang, H.; Hu, Z.; Wei, J.; Xie, P.; Kim, G.; Ho, Q.; and Xing, E. 2015.
\newblock Poseidon: A System Architecture for Efficient GPU-based Deep Learning
  on Multiple Machines.
\newblock In \emph{arXiv preprint 1512.06216}.

\bibitem[{Zhong et~al.(2018)Zhong, Yan, Wu, Shao, and Liu}]{ZhongYWSL18}
Zhong, Z.; Yan, J.; Wu, W.; Shao, J.; and Liu, C.-L. 2018.
\newblock Practical Block-Wise Neural Network Architecture Generation.
\newblock In \emph{CVPR}.

\bibitem[{Zhou et~al.(2020)Zhou, Zhou, Zhang, Loy, Yi, Zhang, and
  Ouyang}]{EcoNAS_CVPR2020}
Zhou, D.; Zhou, X.; Zhang, W.; Loy, C.~C.; Yi, S.; Zhang, X.; and Ouyang, W.
  2020.
\newblock EcoNAS: Finding Proxies for Economical Neural Architecture Search.
\newblock In \emph{CVPR}.

\bibitem[{Zoph and Le(2017)}]{NAS_RL_ICLR2017}
Zoph, B.; and Le, Q.~V. 2017.
\newblock Neural Architecture Search with Reinforcement Learning.
\newblock In \emph{ICLR}.

\bibitem[{Zoph et~al.(2018)Zoph, Vasudevan, Shlens, and Le}]{ZophVSL17}
Zoph, B.; Vasudevan, V.; Shlens, J.; and Le, Q.~V. 2018.
\newblock Learning Transferable Architectures for Scalable Image Recognition.
\newblock In \emph{CVPR}.

\end{thebibliography}

\clearpage
\newpage

%%%%%%%%% Supplementary %%%%%%%%%
\section{Supplementary Material}

\section{A. Datasets} We use the following four commonly used image datasets, and derive the proxy sets from them to use for our transfer experiments.

\noindent \textbf{MIT67}~\cite{quattoni2009recognizing}: This is a scene recognition dataset which contains 15620 images of 67 Indoor categories. The minimum number of images per category is 100 and the dataset is available at \url{http://web.mit.edu/torralba/www/indoor.html}.

\noindent \textbf{FLOWERS102}~\cite{nilsback2008automated}: a collection of flower images of 102 categories that are commonly occurring in the United Kingdom. Each class consists of between 40 and 258 images and the dataset is publicly available at \url{https://www.robots.ox.ac.uk/~vgg/data/flowers/102/}. 

\noindent\textbf{CIFAR10} \cite{Krizhevsky09learningmultiple}: a collection of 50K training and 10K validation images with resolution 32x32, annotated in 10 classes with uniform distribution. Training and validation splits are provided with the dataset which is available to download at \url{https://www.cs.toronto.edu/~kriz/cifar.html}.

\noindent\textbf{CIFAR100} \cite{Krizhevsky09learningmultiple}: the same collection as CIFAR10 of 50K training and 10K validation natural images with resolution 32x32, but annotated in 100 classes with uniform distribution. The dataset is publicly available to download at \url{https://www.cs.toronto.edu/~kriz/cifar.html}.

\noindent\textbf{ImageNet1K} \cite{imagenet_cvpr09}: a collection of 1.2M training and 50K validation natural images with resolution 256x256, annotated in 1,000 classes with uniform distribution. Training and validation splits are provided with the dataset which is available at \url{http://image-net.org/download}.

\begin{figure}[ht]
    \centering
     \includegraphics[trim={15mm 35mm 5mm 30mm},clip, width=0.95\linewidth]{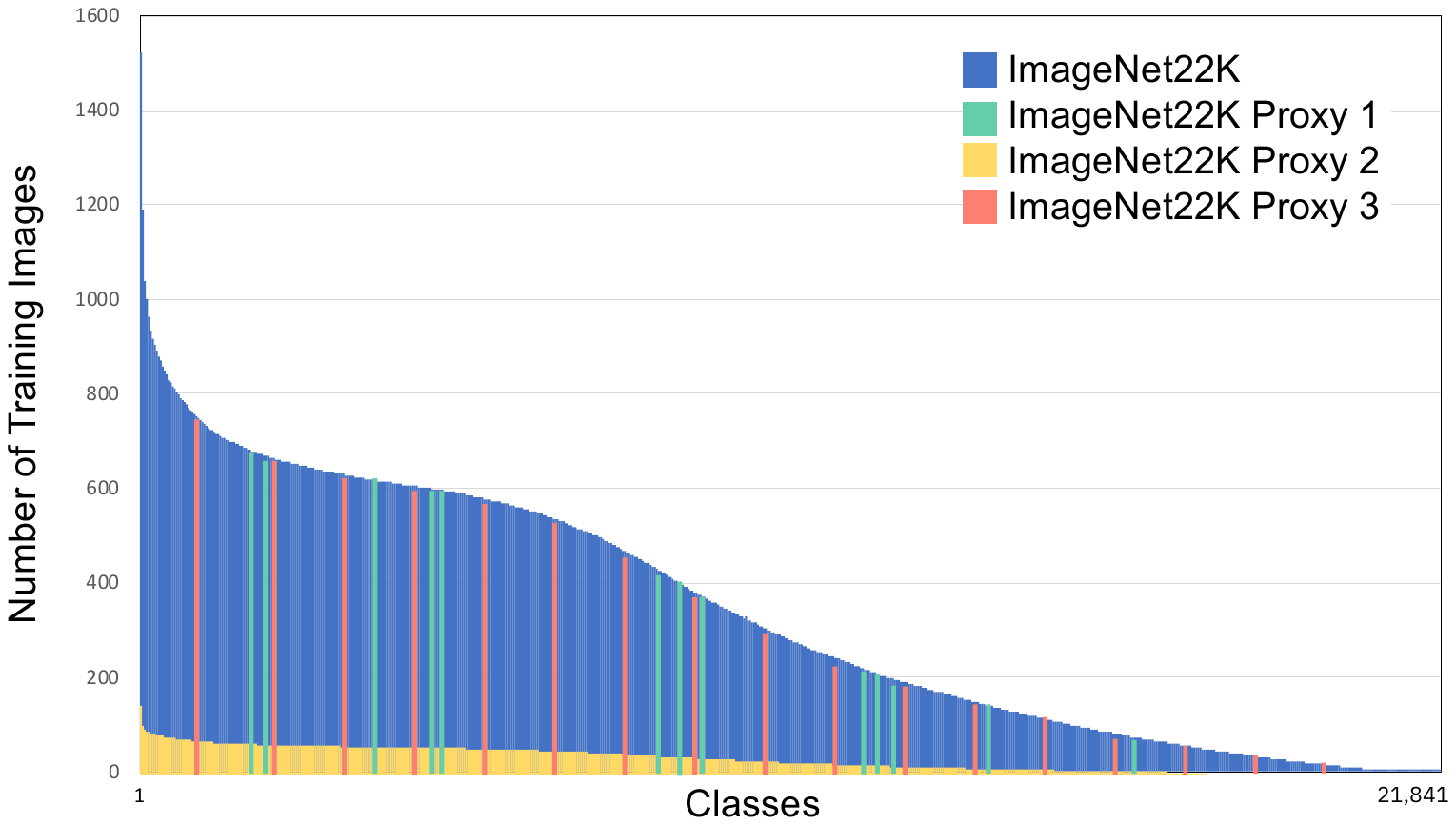}
    \caption{\textbf{ImageNet22K}. Distribution of training images and visual representation of sampling strategies for the Proxy Sets. While the largest class contains 2,469 images, 479 classes have only 1 image. On average, there are 343 images per class. Best viewed in color.}
    \label{fig:imagenet22K}
\end{figure}

\noindent\textbf{ImageNet22K} \cite{ILSVRC15}: a collection of over 14M natural images with resolution 256x256, annotated in 21,841 classes with heavily skewed distribution (as shown in Figure \ref{fig:imagenet22K}, in blue). While the largest class contains 2,469 images, 479 classes have only 1 image. On average, there are 343 images per class. Since there is not an official training/test split for the dataset, in our augment experiments we followed the common practice \cite{adamOSDI2014, cho2017powerai} of randomly splitting the dataset in two halves, one for training and one for testing. We adopted the same split as Cho et al. \cite{cho2017powerai}. This dataset is publicly available to download at \url{http://image-net.org/download}.

\noindent\textbf{Proxy Sets}. Besides CIFAR10 and CIFAR100, we designed a series of Proxy-Sets by sampling classes and images from ImageNet1K and ImageNet22K, as follows:

\noindent\textbf{ImageNet1K Proxy 1}. 100 randomly sampled classes from ImageNet1K.

\noindent\textbf{ImageNet1K Proxy 2}. 200 randomly sampled classes from ImageNet1K.

\noindent\textbf{ImageNet1K Proxy 3}. 300 randomly sampled classes from ImageNet1K.

\noindent\textbf{ImageNet1K Proxy 4}. 200 randomly sampled classes from ImageNet1K, same number of exsamples as Proxy 1.

\noindent\textbf{ImageNet1K Proxy 5}. 1,000 classes from ImageNet1K, same number of exsamples as Proxy 1.

\noindent\textbf{ImageNet22K Proxy 1}. 100 randomly sampled classes from ImageNet22K, as illustrated in in Figure \ref{fig:imagenet22K}, in green.

\noindent\textbf{ImageNet22K Proxy 2}. 13,377 classes, randomly sampled images from each class of ImageNet22K to obtain same cumulative number of examples as Proxy 1 (Fig.~\ref{fig:imagenet22K}, in yellow).

\noindent\textbf{ImageNet22K Proxy 3}. 100 uniformly sampled classes from ImageNet22K, to obtain same cumulative number of examples as Proxy 1 (Figure~\ref{fig:imagenet22K}, in red).

\begin{table*}[t]
\resizebox{\textwidth}{!}{%
\begin{tabular}{c|c|c|c|c|c|c|c|c}
\toprule
Search & \multirow{2}{*}{CIFAR10} & \multirow{2}{*}{CIFAR100} &  ImageNet1K  & ImageNet1K  & ImageNet1K  & ImageNet22K  & \multirow{2}{*}{ImageNet1K} & \multirow{2}{*}{ImageNet22K} \\ 
Method & & & Proxy 1 & Proxy 2 & Proxy 3 & Proxy 1 & &  \\ \hline
ENAS  & 0.375 & 0.375 & 1.29 & 2.21 & 3 & 1 & 12 & 414 \\ %\hline
DARTS & 0.458 & 0.509 & 2.65 & 5.30 & 7.95 & 1.08 & 25 & 447  \\ %\hline
NAO   & 2.14 & 2.16 & 3.38 & 6.32 & 9.62 & 1.35 & 33 & 560  \\ 
NSGANET   & 4 & 4 & 60 & 121 & 180 & 16 & 600 & 6,628    \\ 
\bottomrule
\end{tabular}%
}
\caption{Search times (in GPU-days) for each method on the proxy sets (measured) and on the direct sets (estimates). NSGANet requires maximum time for search among all the methods.}
\label{tab:search_time}
\end{table*}

\section{B. Training Protocol Details}

In all our experiments we use the codebase\footnote{\url{https://github.com/antoyang/NAS-Benchmark}} from Yang et al. \cite{Yang2020NAS} running Pytorch on NVIDIA Tesla V100 GPUs with 16GB memory. In the following we describe in detail the parameters of the search protocols for each method.
\textbf{DARTS:} Batch size = 64, SGD optimizer, weight decay=3e-4, momentum = 0.9, seed = 2, lr = 0.025; 
\textbf{ENAS:} Batch size = 128, SGD optimizr, seed = 2, lr = 0.05, momentum = 0.9, weight decay = 1e-4, controller optimizer = Adam;
\textbf{NSGANet:} Batch size = 128, SGD optimizer, seed = 2, lr = 0.025, weight decay = 3e-4, momentum = 0.9, number of offspring is 20, number of generations is 30, population size is 40;
\textbf{NAO:} Batch size = 64, SGD optimizer, seed = 2, lr = 0.1, learning rate decay = 0.97, momentum = 0.9.
For augmentations, batch size was fixed at 128 per GPU across all augment runs, except for models using 40 and 60 cells, where the memory constraints required smaller batch sizes of 96 and 64 per GPU, respectively.
For augmentations we followed the default settings as the experiments in \cite{Yang2020NAS}, using cross entropy loss, SGD optimizer with learning rate 0.025, momentum 0.9, seed 2, initial number of channels 36, and gradient clipping set at 5. 
Default DAC training protocol values were Auxiliary Towers weight $A = 0.4$, Cutout Length $C = 16$ and Drop-path Probability $D = 0.2$. In Table \ref{tab:search_time} we report the single GPU search times for each method, measured in GPU-days for all the small and proxy sets, and estimated for ImageNet1K and ImageNet22K based on those measurements.

\begin{figure*}[h!]
    \centering
    %\captionsetup[subfigure]{labelformat=empty}
    \begin{subfigure}{0.3\linewidth}
        \includegraphics[width=\linewidth]{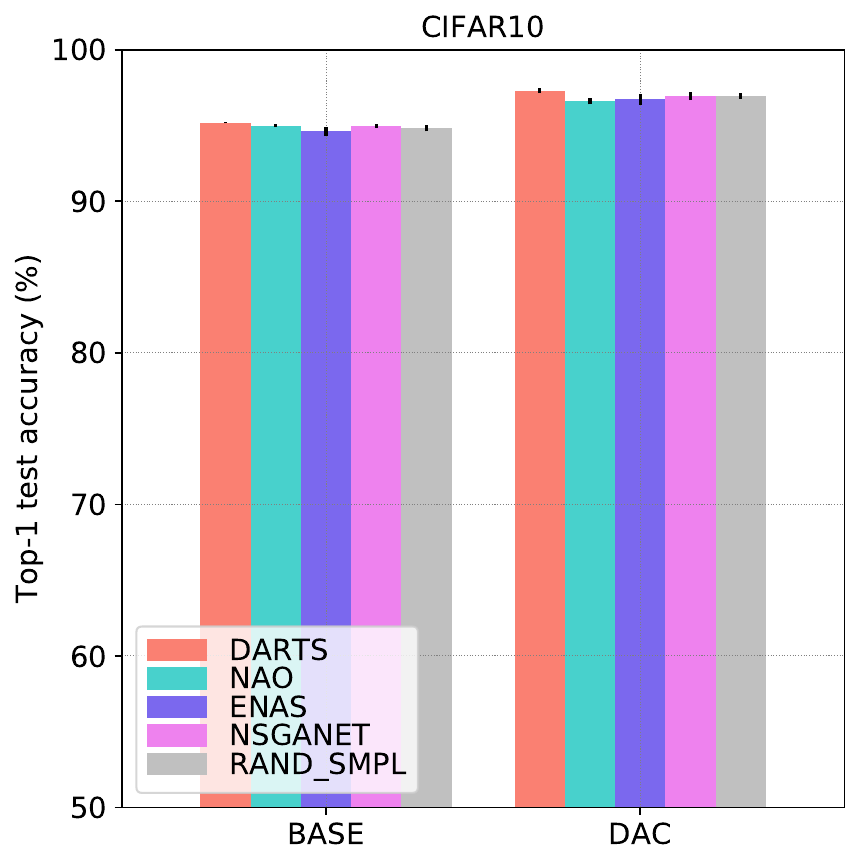}
        \label{fig:transfer_vs_direct_exp1_cifar10}
    \end{subfigure}%
    \begin{subfigure}{0.3\linewidth}
        \includegraphics[width=\linewidth]{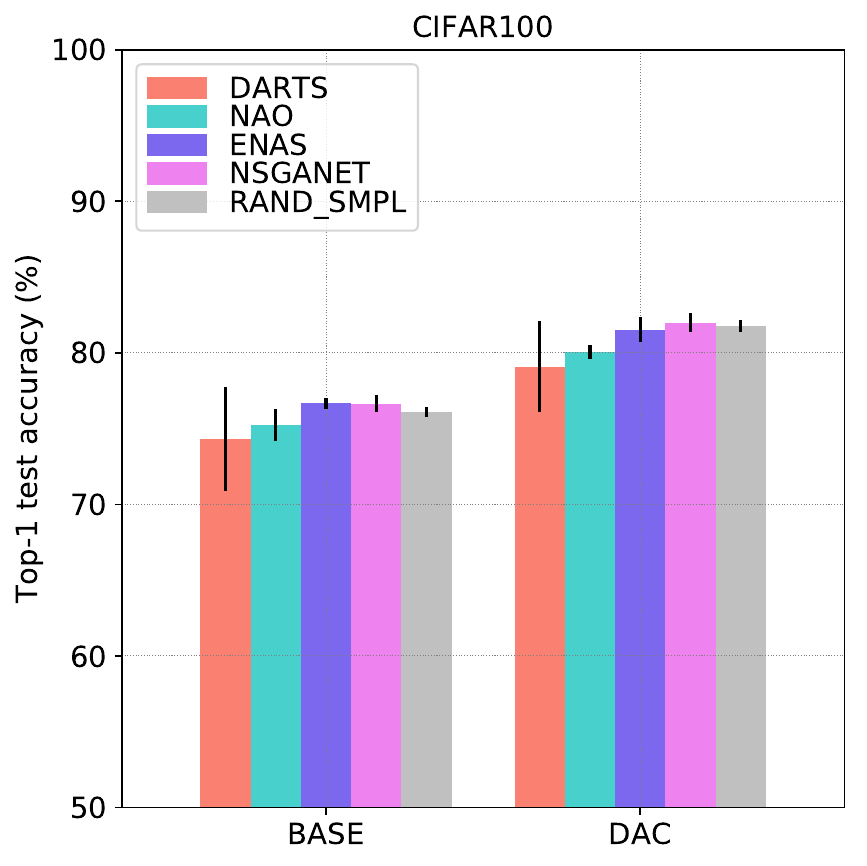}
        \label{fig:transfer_vs_direct_exp1_cifar100}
    \end{subfigure}%
    \begin{subfigure}{0.3\linewidth}
        \includegraphics[width=\linewidth]{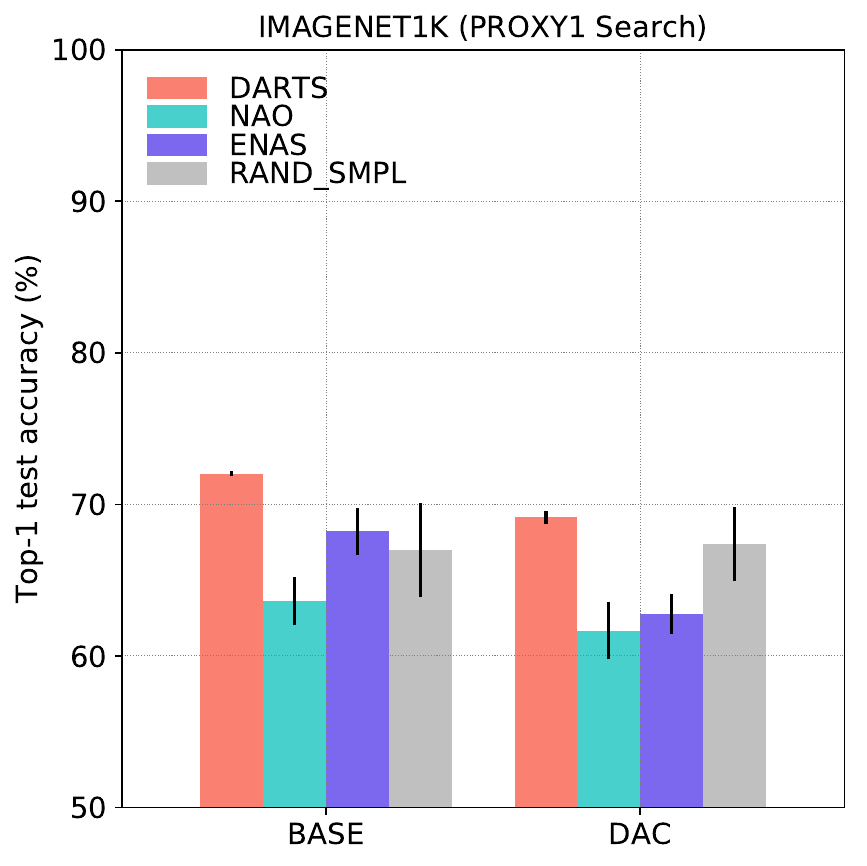}
        \label{fig:transfer_vs_direct_exp1_im1k}
    \end{subfigure}%
    \vspace{-4mm}
    \caption{\textbf{Effect of training protocol on direct experiments}. 
    %(a) CIFAR10, (b) CIFAR100, (c) ImageNet1K. 
    DAC training protocol improves the performance of different NAS methods on CIFAR10 and CIFAR100, but its importance is surprisingly negligible on ImageNet1K direct experiments. Best viewed in color.}
    \label{fig:transfer_vs_direct}
\end{figure*}

\section{Effect of Training Protocol}
\begin{figure*}[ht]
    \centering
    %\captionsetup[subfigure]{labelformat=empty}
    \begin{subfigure}{0.25\linewidth}
        \includegraphics[width=\linewidth]{fig/4_base_vs_dac_direct_IMAGENET1K_PROXY1.pdf}
        %\caption{}
        \label{fig:transfer_vs_direct_exp1_im1kp1}
    \end{subfigure}%
    \begin{subfigure}{0.25\linewidth}
        \includegraphics[width=\linewidth]{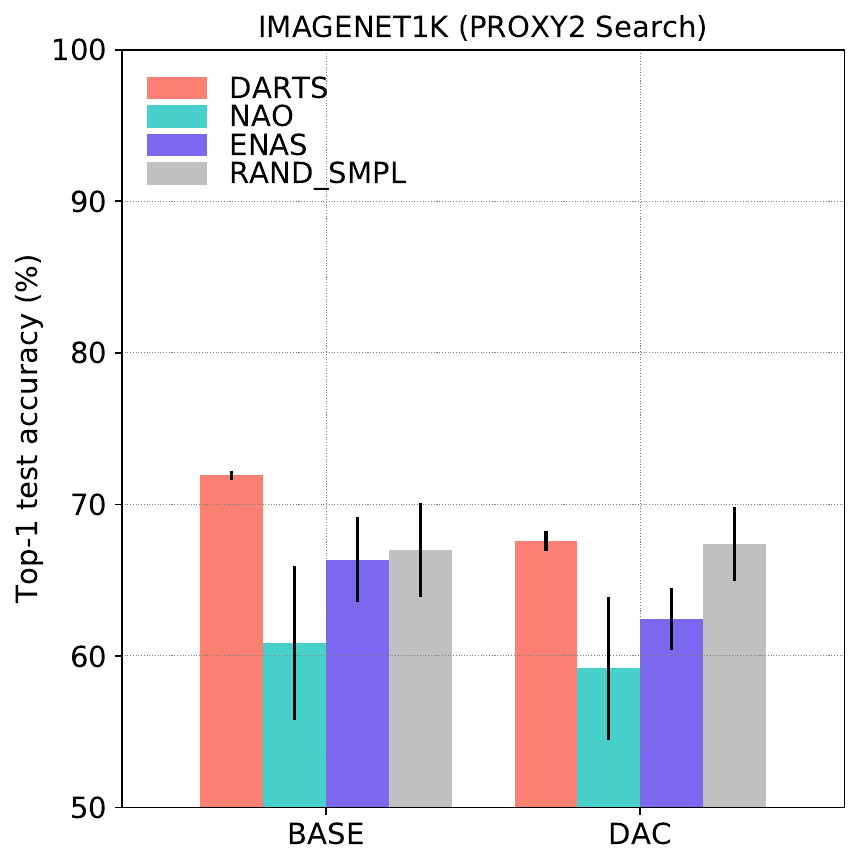}
        %\caption{}
        \label{fig:transfer_vs_direct_exp1_im1kp2}
    \end{subfigure}%
    \begin{subfigure}{0.25\linewidth}
        \includegraphics[width=\linewidth]{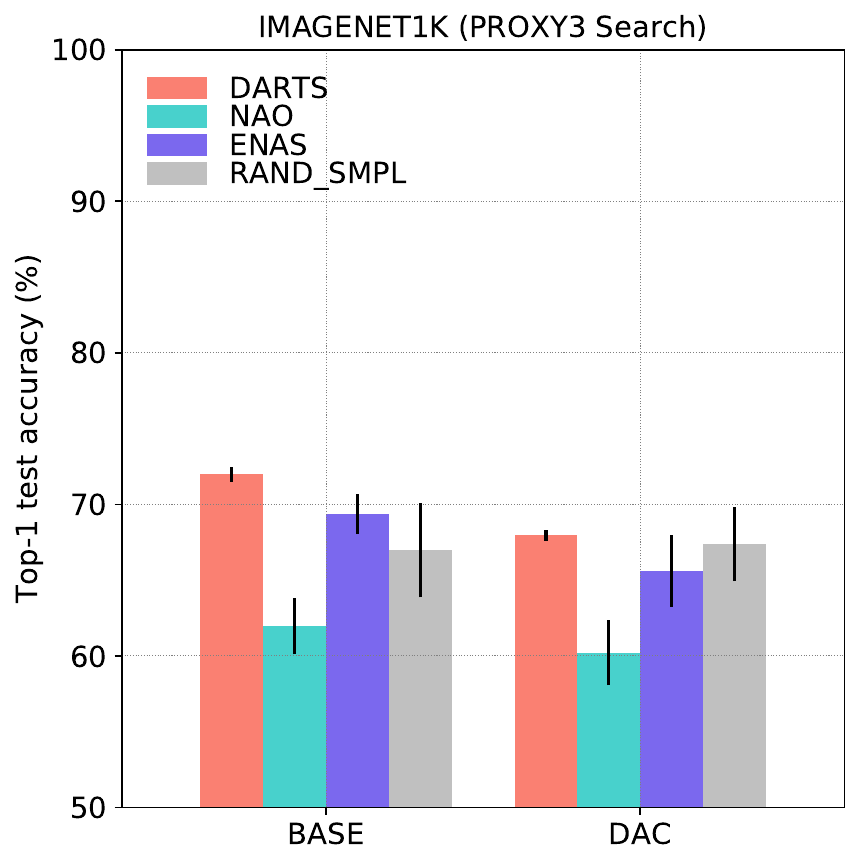}
        %\caption{}
        \label{fig:transfer_vs_direct_exp1_im1kp3}
    \end{subfigure}%
    \begin{subfigure}{0.25\linewidth}
        \includegraphics[width=\linewidth]{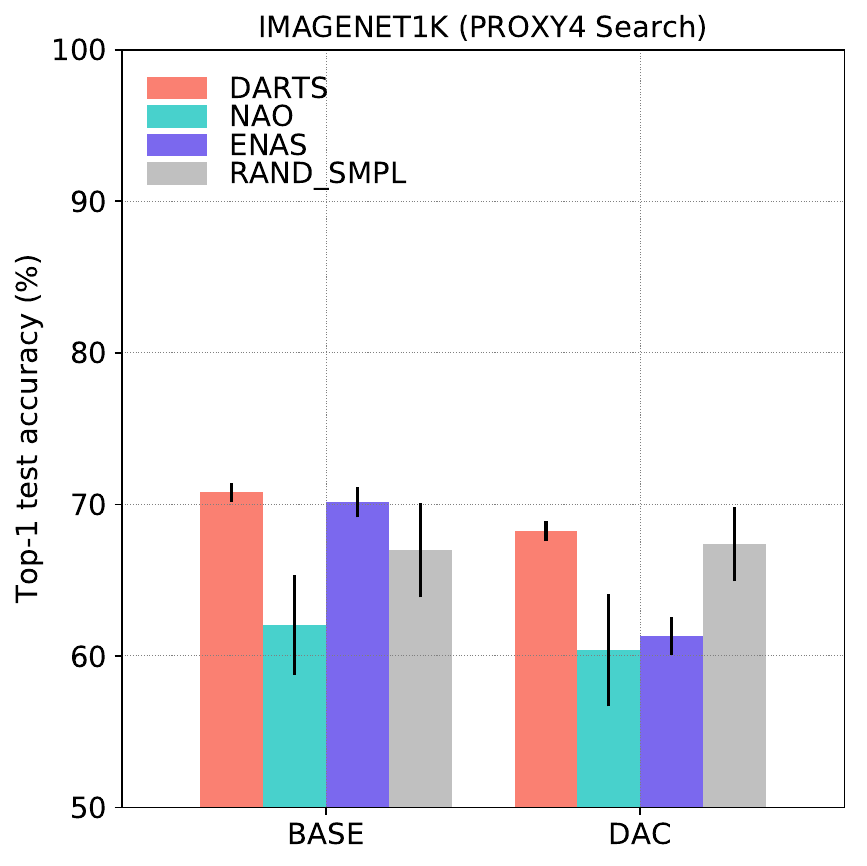}
        %\caption{}
        \label{fig:transfer_vs_direct_exp1_im1kp4}
    \end{subfigure}

    \begin{subfigure}{0.25\linewidth}
        \includegraphics[width=\linewidth]{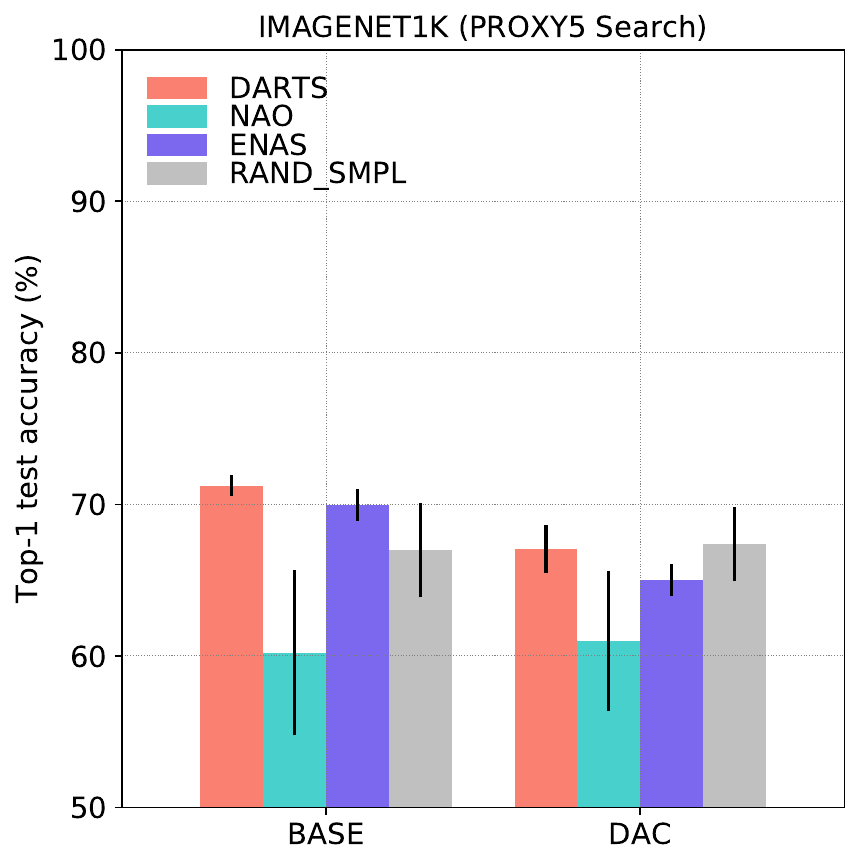}
        %\caption{}
        \label{fig:transfer_vs_direct_exp1_im1kp5}
    \end{subfigure}%
    \begin{subfigure}{0.25\linewidth}
        \includegraphics[width=\linewidth]{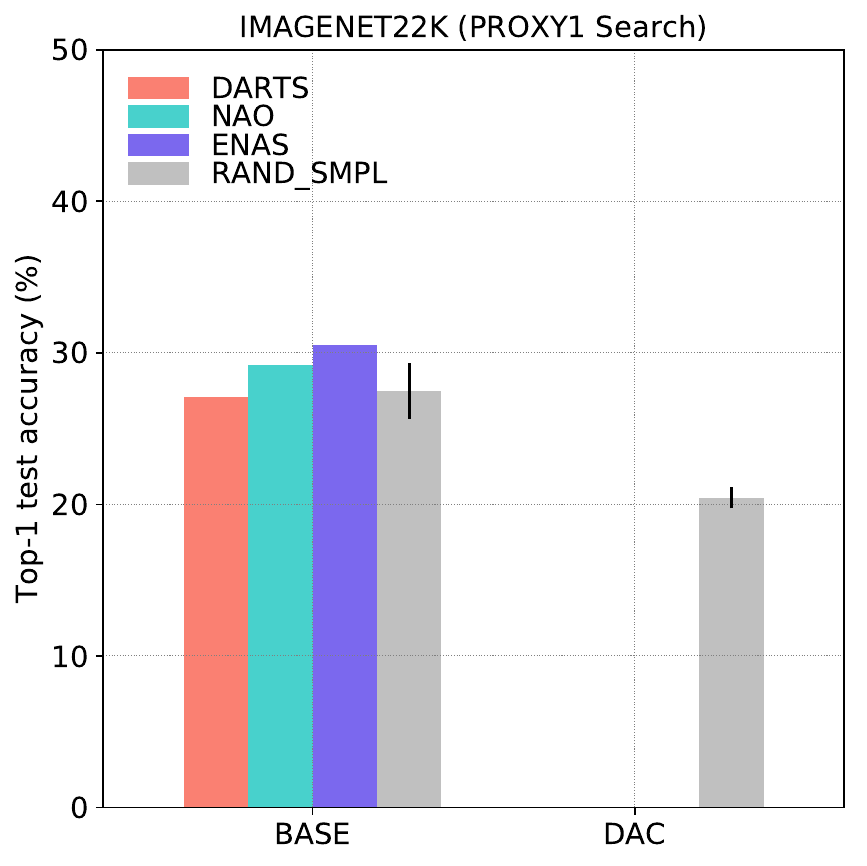}
        %\caption{}
        \label{fig:transfer_vs_direct_exp1_im22kp1}
    \end{subfigure}%
    \begin{subfigure}{0.25\linewidth}
        \includegraphics[width=\linewidth]{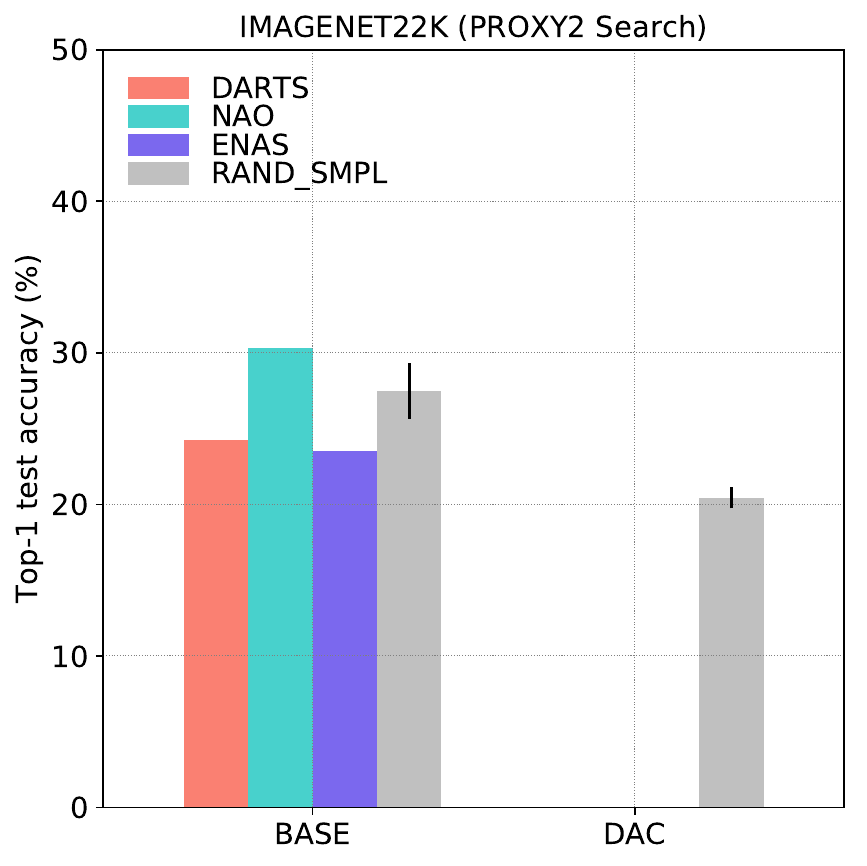}
        %\caption{}
        \label{fig:transfer_vs_direct_exp1_im22kp2}
    \end{subfigure}%
    \begin{subfigure}{0.25\linewidth}
        \includegraphics[width=\linewidth]{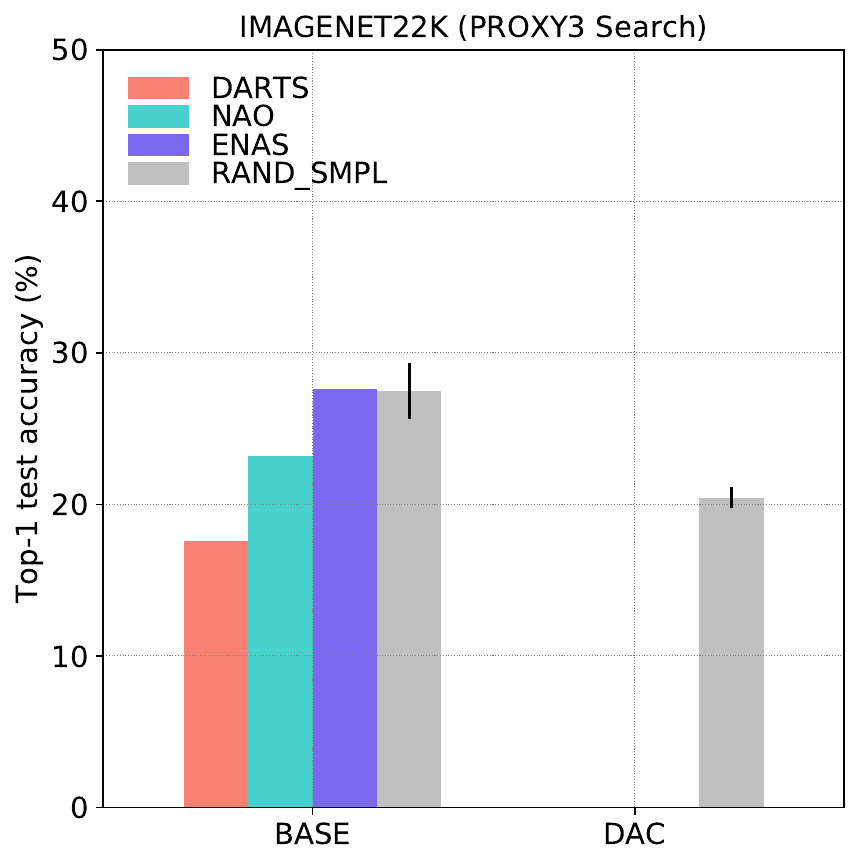}
        %\caption{}
        \label{fig:transfer_vs_direct_exp1_im22kp3}
    \end{subfigure}%
       \vspace{-0.5cm}
    \caption{\textbf{Effect of training protocol on direct experiments}. BASE and DAC augmentation strategies on different proxy sets of ImageNet1K. NAS methods were only trained with BASE augmentation for ImageNet22K.}
    \label{fig:transfer_vs_direct_exp1}
\end{figure*}

Techniques such as Drop path \cite{larsson2017fractalnet}, Auxiliary towers \cite{randomly_wired_ICCV2019} and Cutout \cite{devries2017cutout} have been developed as general learning tools to improve the accuracy of deep learning models in different applications. As such, they should be dataset and architecture agnostic. When analyzing our augmentation training results in Figure \ref{fig:transfer_vs_direct} and \ref{fig:transfer_vs_direct_exp1}, we observe that the DAC training protocol provides significant benefits in terms of Top-1 accuracy for the smaller datasets (CIFAR10 and CIFAR100), confirming the observations of Yang et al. \cite{Yang2020NAS}.
However, its importance becomes surprisingly negligible on the larger datasets. For example on ImageNet1K (see Figure \ref{fig:transfer_vs_direct_exp1} - ImageNet1K), BASE protocol achieves an accuracy of $72.03\pm0.2$ for DARTS versus $63.64\pm1.78$ for NAO, whereas the DAC protocol achieves an accuracy of only $69.13\pm0.45$ and $61.67\pm2.11$ for DARTS and NAO respectively. 
The same behavior can also be observed in Figure \ref{fig:transfer_vs_direct_exp2} for the transfer results. Even when transferring, training protocol (DAC methods) are more important for performance gains than search methods or proxy sets adopted for small scale datasets (CIFAR100). When looking at larger datasets such as ImageNet, choices of search methods and proxy sets adopted become more important. For example picking CIFAR10 as proxy set over CIFAR100 for DARTS yields an absolute improvement on ImageNet1K of $7.93\%$ ($71.3\pm0.78$ versus $63.37\pm4.99$, respectively). The difference between BASE and DAC protocols for almost all search methods on ImageNet1K is instead less than $4\%$ when using either CIFAR10 or CIFA100 as proxy, with the only exception of NSGANet, which exhibits a very wide standard deviation in DAC based runs.
Interestingly this result suggests that the choice of an appropriate proxy set and search strategy is more relevant than some of the ``tricks'' commonly used to improve final classification performance when analyzing NAS at large scale.

\section{Additional Transfer Results}
\begin{figure}[ht]
    \centering
     \includegraphics[width=0.7\linewidth]{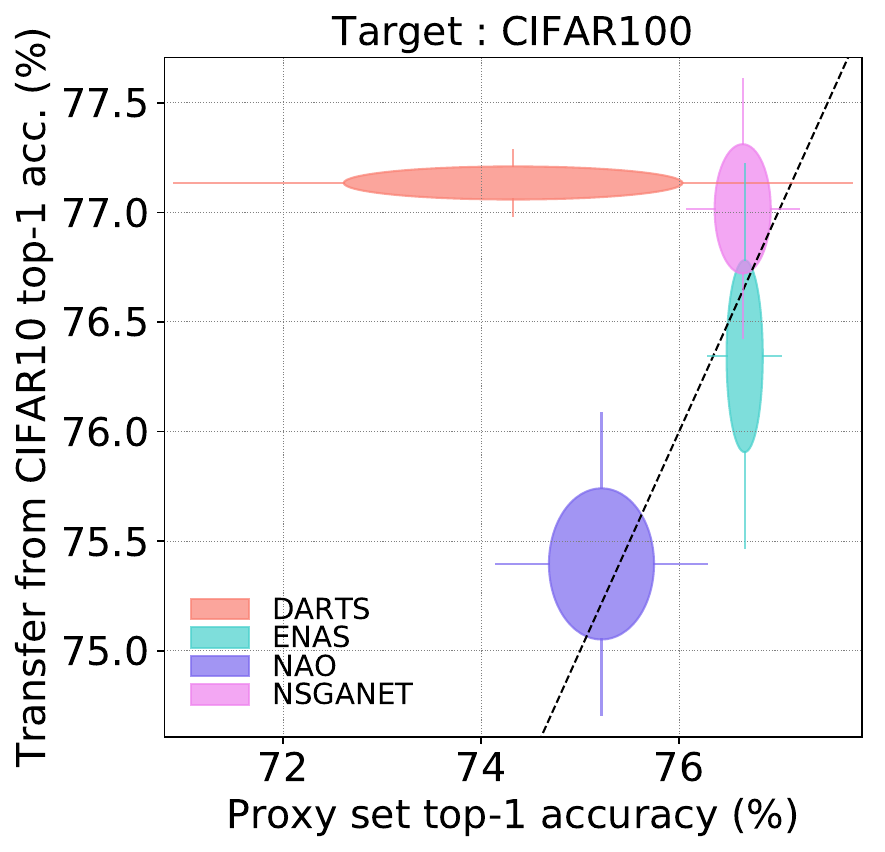}
    \caption{\textbf{Architecture transfer vs Proxy-based direct search}. Comparing architectures transferred from CIFAR10 to those obtained directly from CIFAR100. The transferred architectures perform at-par or better than direct searched architectures across NAS methods.}
    \label{fig:transfer_vs_direct_exp_supp}
\end{figure}

\begin{figure}[ht]
    \centering
    %\captionsetup[subfigure]{labelformat=empty}
    \begin{subfigure}{0.49\linewidth}
    \includegraphics[width=\linewidth]{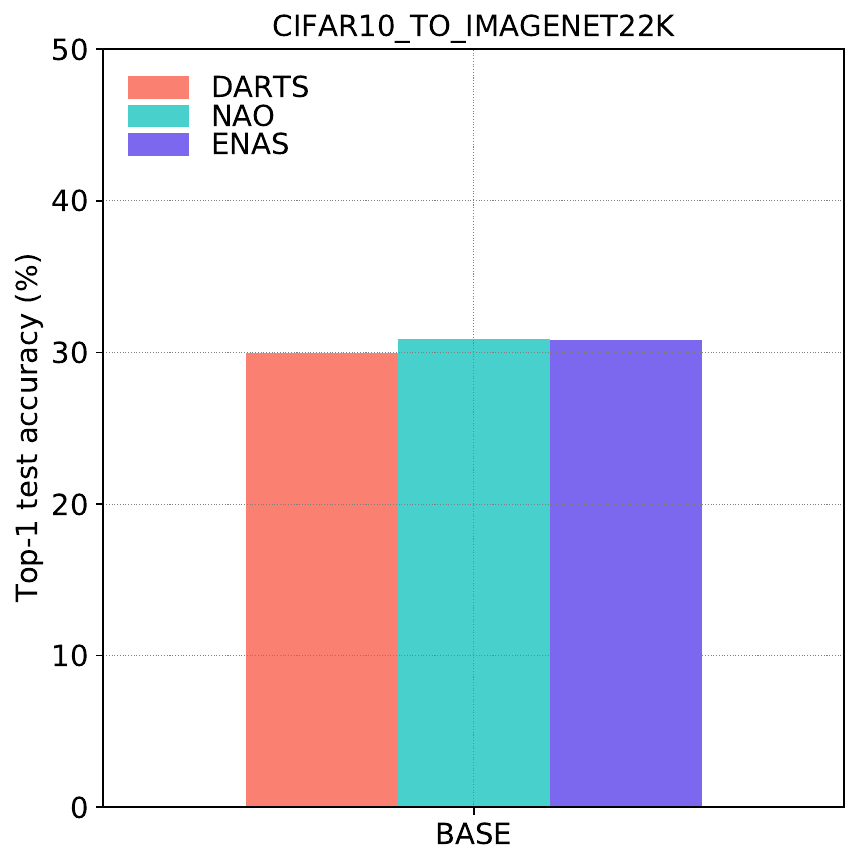}
    \end{subfigure}%
    \begin{subfigure}{0.49\linewidth}
    \includegraphics[width=\linewidth]{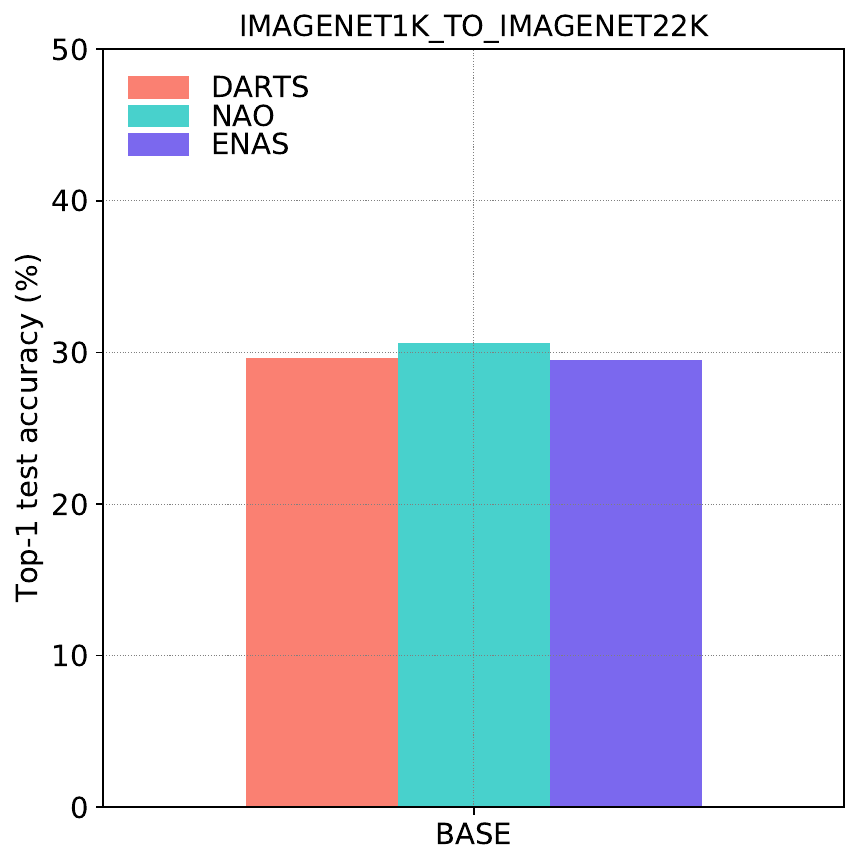}
    \end{subfigure}% 
    \caption{\textbf{Effect of BASE training protocol on large scale transfer experiments}.Transfer from CIFAR10 and ImageNet1K to ImageNet22K.}
    \label{fig:transfer_vs_direct_exp2}
\end{figure}

Figure~\ref{fig:transfer_vs_direct_exp1} shows the effect of training protocol on different proxy sets. As discussed above, the effect of DAC training protocol becomes surprisingly negligible on the larger datasets irrespective of the proxy sets used for searching the architectures.
Figure~\ref{fig:transfer_vs_direct_exp_supp} shows the comparison between architecture transfer and proxy-based direct search on CIFAR100. As can be seen, transfer performance of architectures searched using CIFAR10 can perform similarly or even better than architectures searched directly on proxy target datasets. Figure~\ref{fig:transfer_vs_direct_exp2} shows the effect of base training protocol on transfer experiments on ImageNet22K.

\section{Backward Transfer Results}

\begin{figure} [ht]
  \centering
  %\captionsetup[subfigure]
    \begin{subfigure}{0.5\linewidth}
        \includegraphics[width=\linewidth]{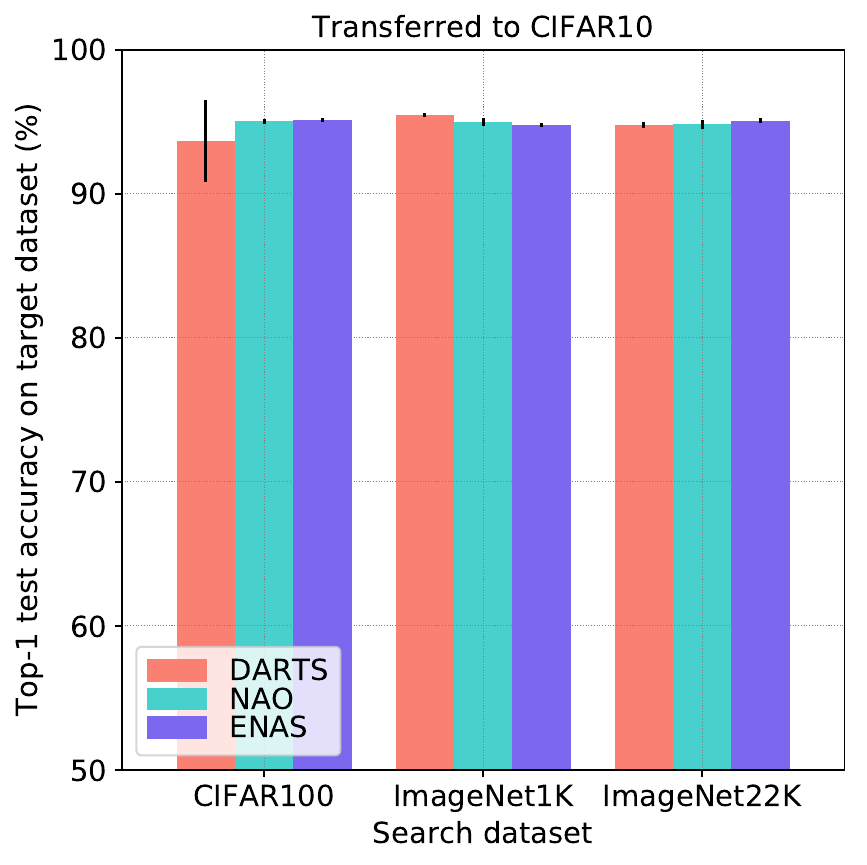}
    \end{subfigure}%
    \begin{subfigure}{0.5\linewidth}
    \includegraphics[width=\linewidth]{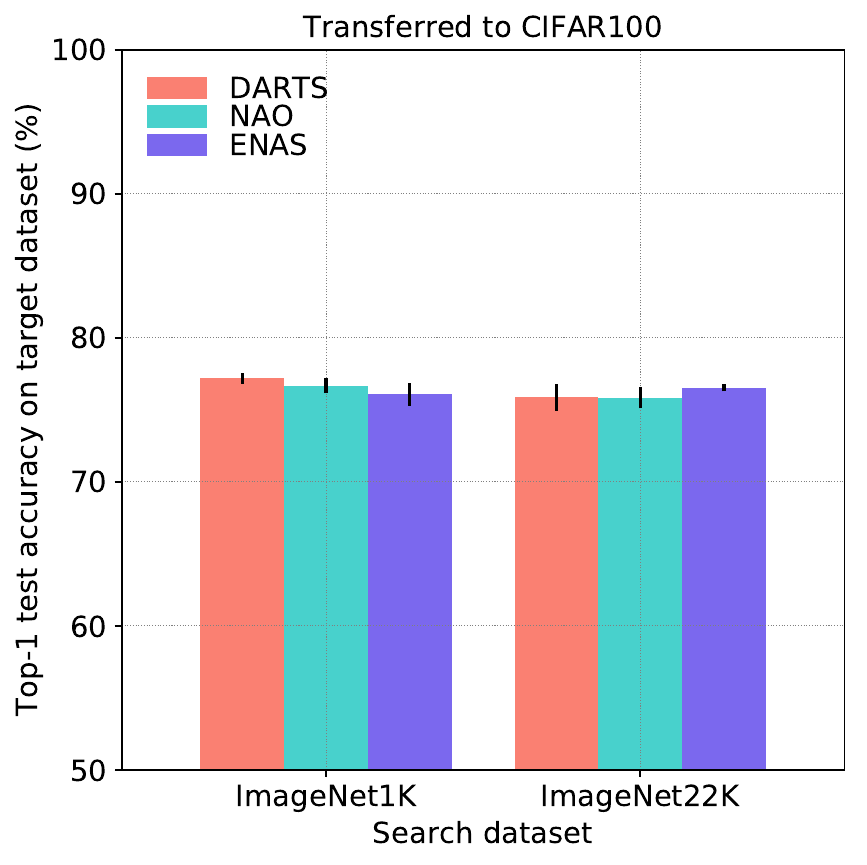}
    \end{subfigure}%
  \caption{\textbf{Architecture transfer performance on small datasets}. Architectures are transferred from CIFAR100 or proxy sets of ImageNet1K or ImageNet22K.}
  \label{fig:transfer_exp_reverse}
\end{figure}

\begin{figure*}[t]
    \centering
    %\captionsetup[subfigure]{labelformat=empty}
    \begin{subfigure}{0.27\linewidth}
        \includegraphics[width=\linewidth]{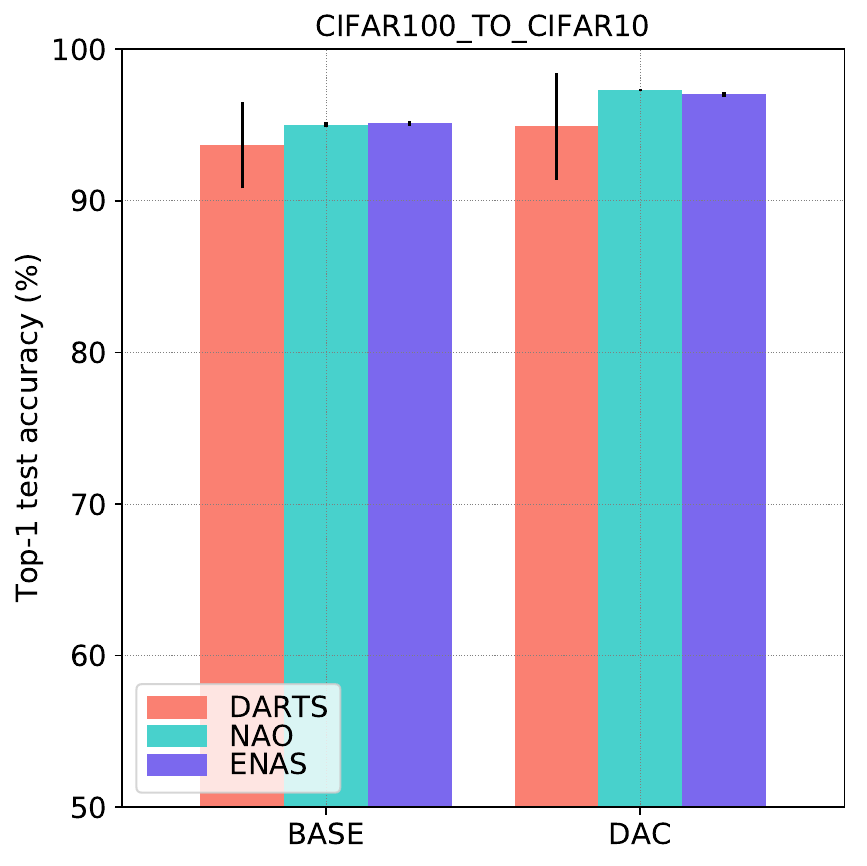}
        %\caption{}
        %\label{fig:transfer_vs_direct_exp2_c100_to_c10}
    \end{subfigure}%
    \begin{subfigure}{0.27\linewidth}
        \includegraphics[width=\linewidth]{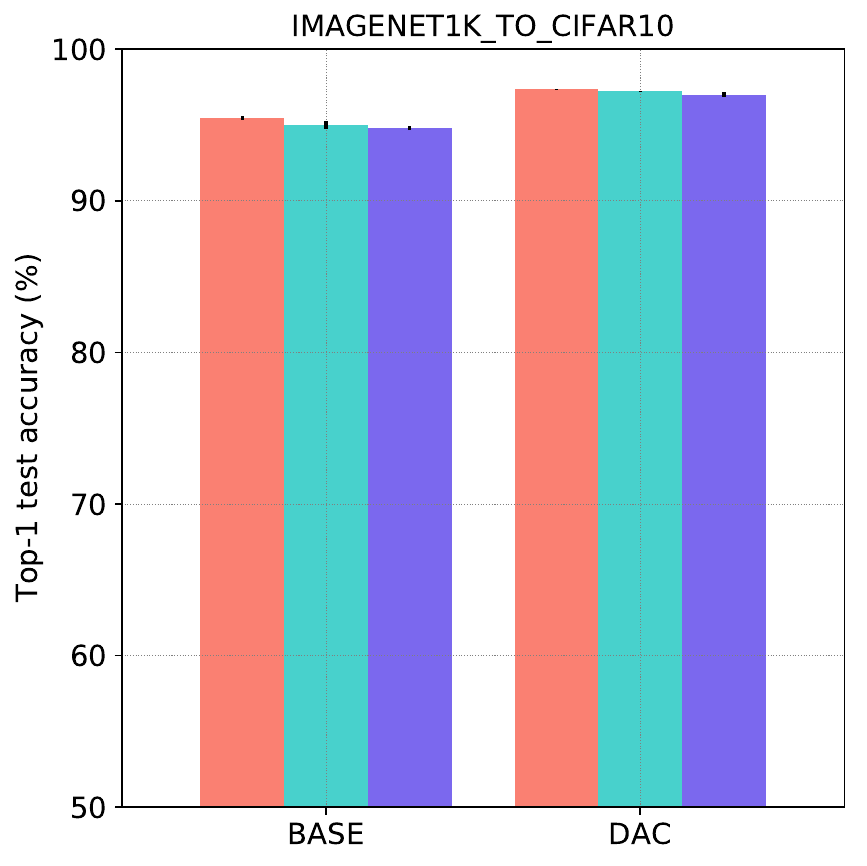}
         %\caption{}
        %\label{fig:transfer_vs_direct_exp2_im1k_to_c10}
    \end{subfigure}%
    \begin{subfigure}{0.27\linewidth}
        \includegraphics[width=\linewidth]{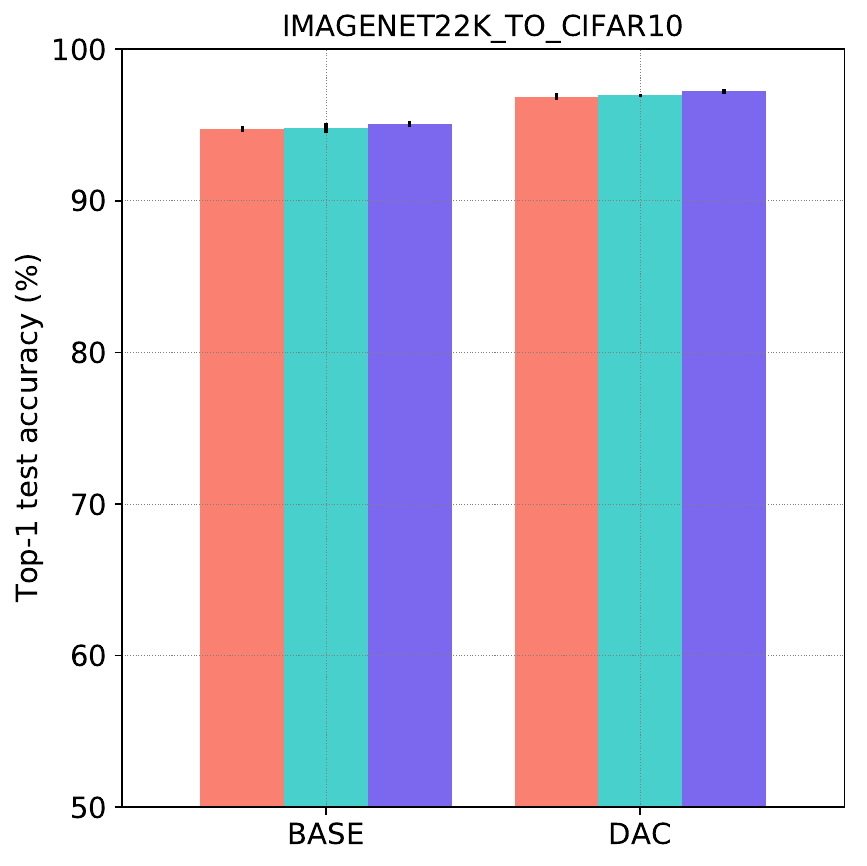}
        
         %\caption{}
        %\label{fig:transfer_vs_direct_exp2_im22k_to_c10}
    \end{subfigure}%
    \\
    \begin{subfigure}{0.27\linewidth}
        \includegraphics[width=\linewidth]{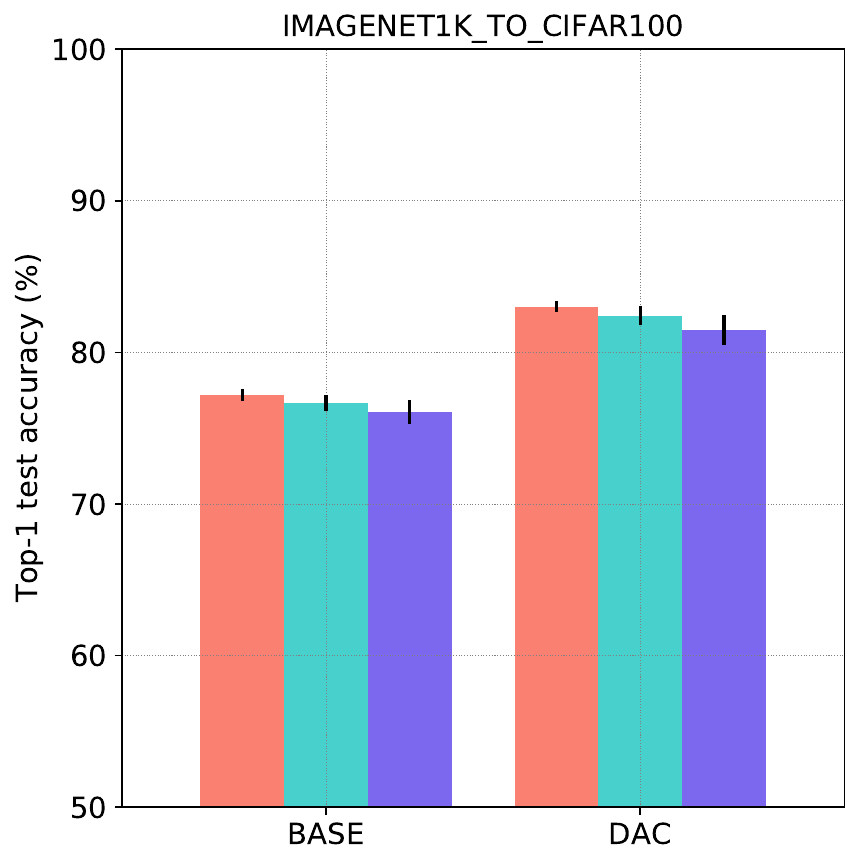}
        %\caption{}
        %\label{fig:transfer_vs_direct_exp2_im1k_to_c100}
    \end{subfigure}%
    \begin{subfigure}{0.27\linewidth}
        \includegraphics[width=\linewidth]{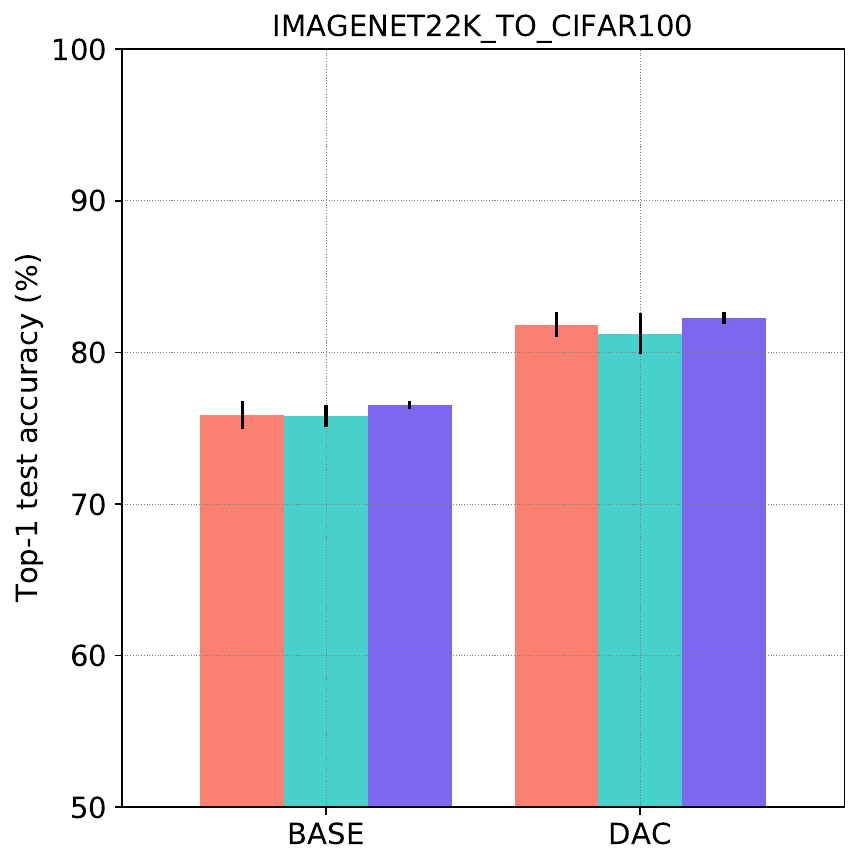}
        %\caption{}
        %\label{fig:transfer_vs_direct_exp2_im22k_to_c100}
    \end{subfigure}% 

    \caption{\textbf{Effect of training protocol on reverse transfer experiments}. Training with DAC augmentation improved test performance across all experiments. Best viewed in color.}
    \label{fig:transfer_vs_direct_exp2_reverse}
\end{figure*}

\begin{figure*}[t]
    \centering
   % \captionsetup[subfigure]{labelformat=empty}
    \begin{subfigure}{0.25\linewidth}
        \includegraphics[width=\linewidth]{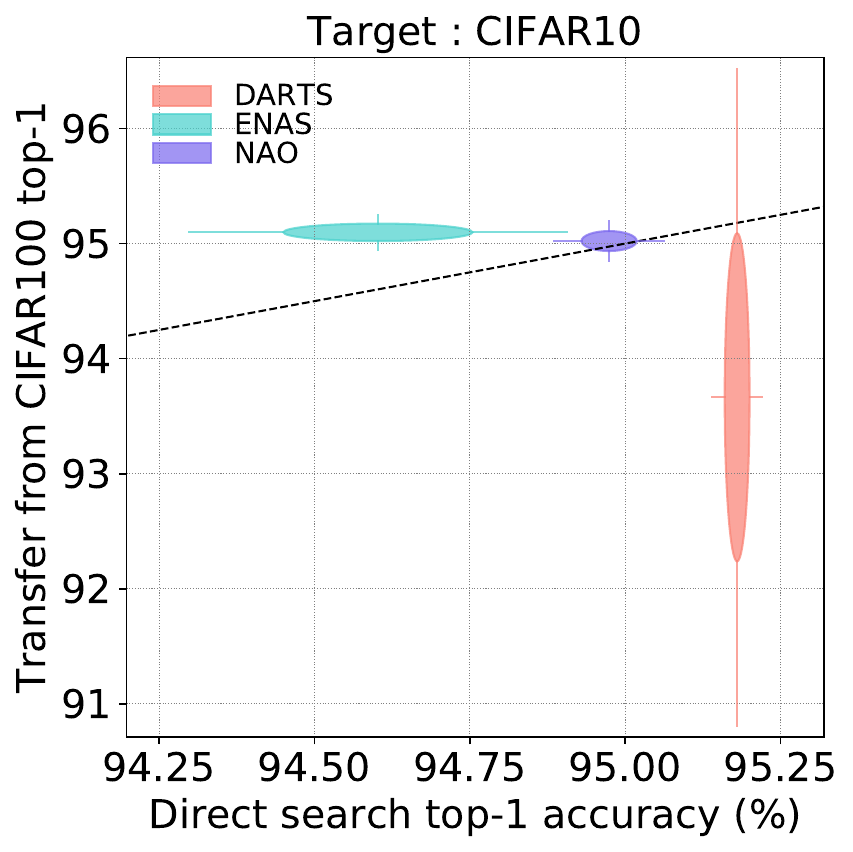}
    \end{subfigure}%
    \begin{subfigure}{0.27\linewidth}
    \includegraphics[width=\linewidth]{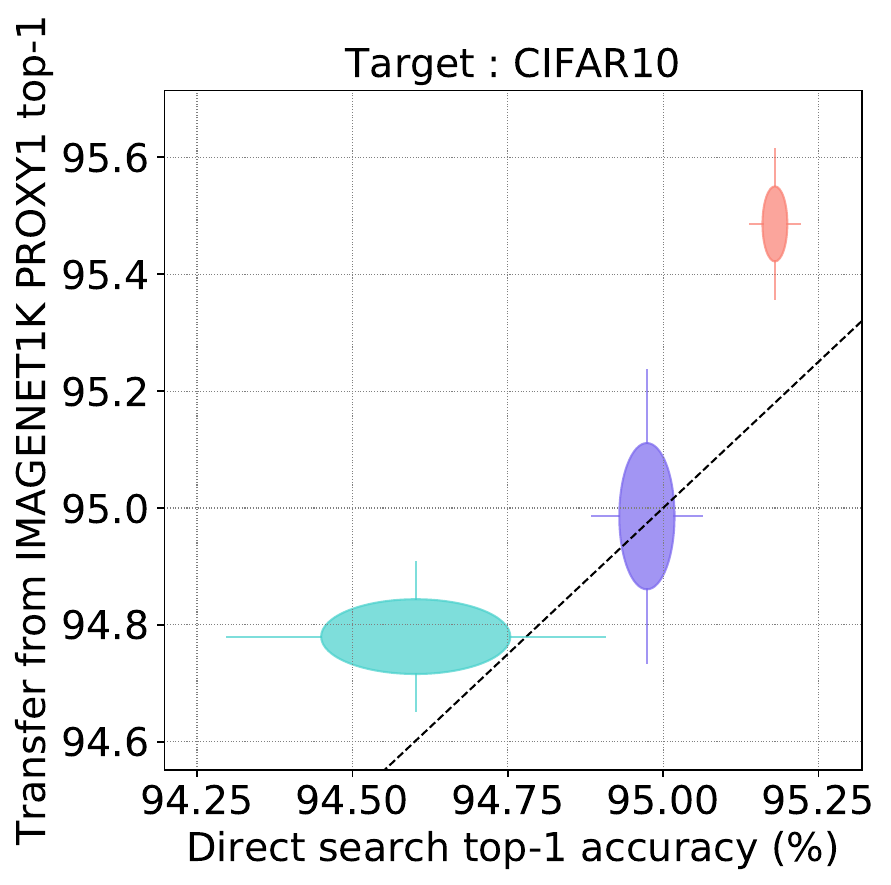}
    \end{subfigure}%
    \begin{subfigure}{0.27\linewidth}
    \includegraphics[width=\linewidth]{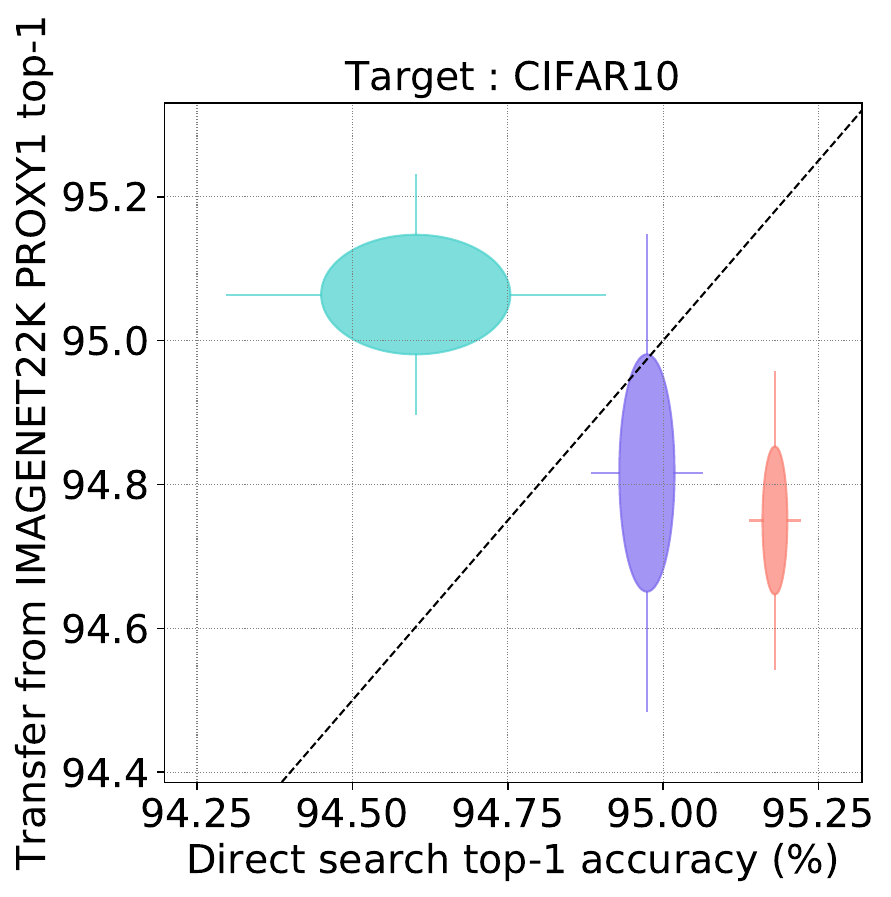}
    \end{subfigure}%
    \\
    \begin{subfigure}{0.27\linewidth}
    \includegraphics[width=\linewidth]{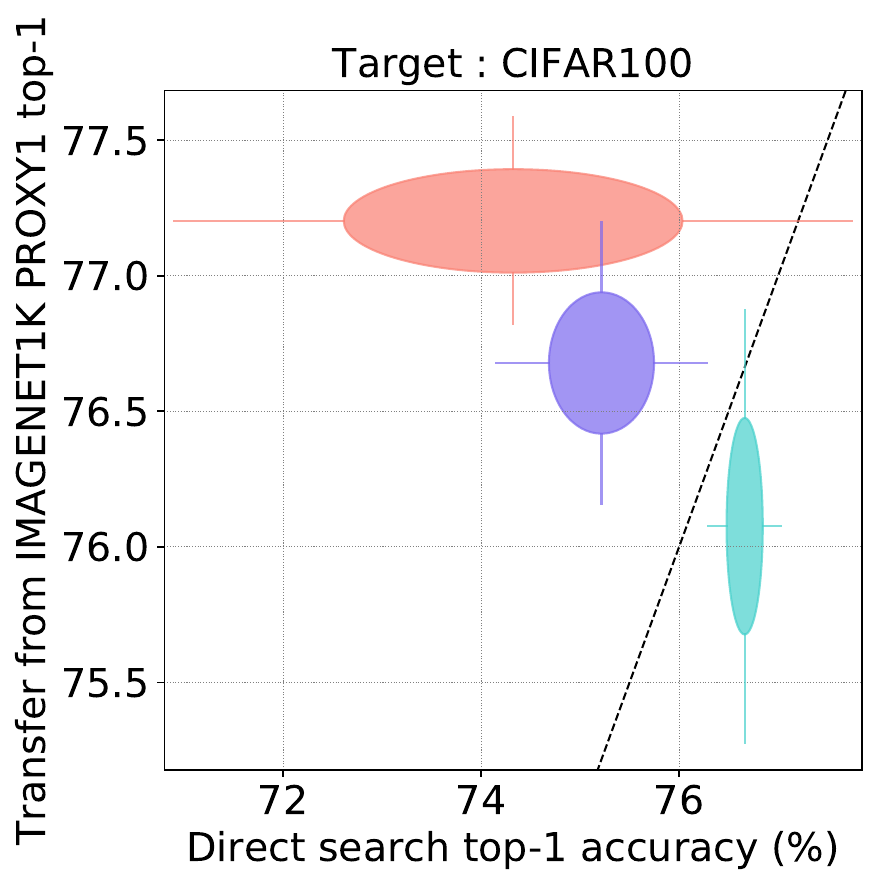}
    \end{subfigure}%
    \begin{subfigure}{0.27\linewidth}
    \includegraphics[width=\linewidth]{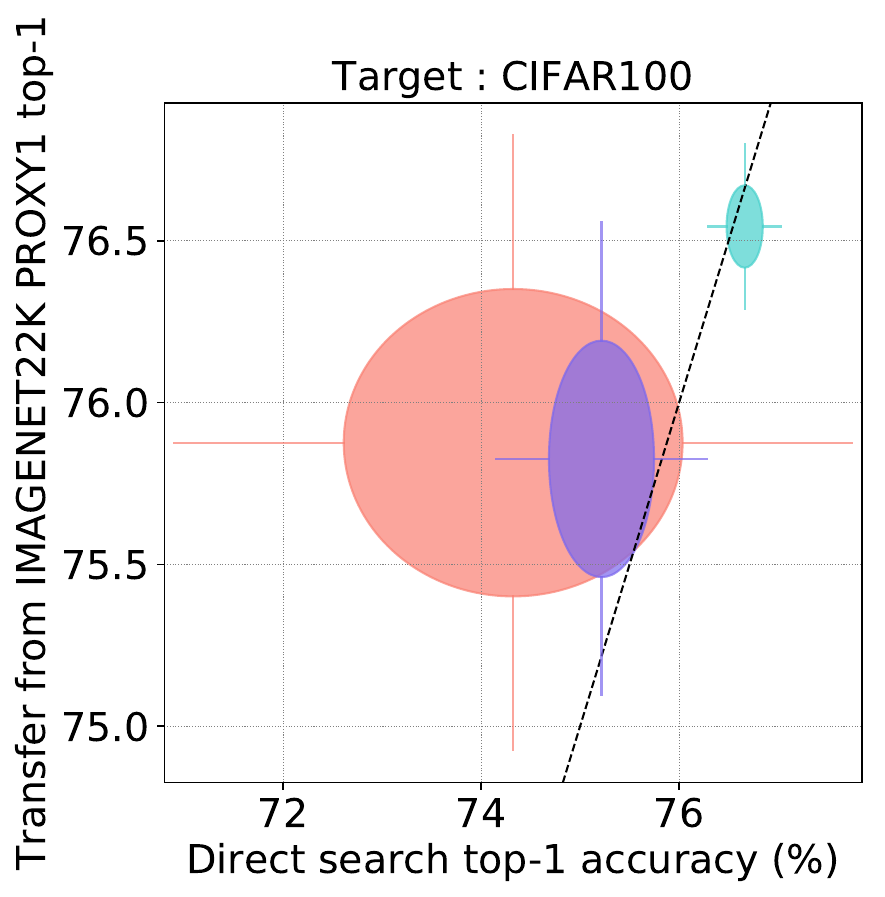}
    \end{subfigure}%
    \caption{\textbf{Architecture transfer vs Direct search}. Comparison of reverse transfer performance with direct search on CIFAR10 and CIFAR100 datasets. Methods lying in the diagonal indicate that transfer performance is similar to the direct search, while methods above the diagonal outperform it.}
    \label{fig:transfer_vs_direct_exp_reverse}
\end{figure*}

In the main paper, we presented transfer results for searching on smaller-scale datasets and transferring the architectures to larger-scale datasets, such as ImageNet1K and ImageNet22K. In this section, we present results where the architectures were searched either on proxy sets of large-scale datasets (e.g.: ImageNet1K proxy set1) or on whole datasets (e.g.: CIFAR100) and transferred to smaller datasets such as CIFAR10 or CIFAR100. Our goal is to see if the cells found with higher resolution is useful when transferred to lower resolution images. We conducted analysis of backward transferring architectures akin to the experiments conducted in the main paper.
Figure~\ref{fig:transfer_exp_reverse} shows the Top-1 test accuracy of transferring architectures searched on large datasets to smaller datasets, such as CIFAR10 and CIFAR100. DARTS transferred from CIFAR100 to CIFAR10 seems to have the largest standard deviation of test accuracy. Figure~\ref{fig:transfer_vs_direct_exp_reverse} compares tranferring from larger datasets to directly searching on small-scale datasets. For CIFAR10, transferring from larger datasets seems to helpful for ENAS, but for NAO and DARTS performance depends upon the type of large dataset. For CIFAR100, transferring from ImageNet datasets boosts performance of DARTS and NAO, while ENAS performes best with direct search. Overall, searching on data that has larger number of images per class seems to help tranfer performance. The effect of training the reverse transferred architectures with BASE and DAC augmentation strategies are illustrated in Figure~\ref{fig:transfer_vs_direct_exp2_reverse}. Using DAC augmentation improved top-1 test accuracy for all NAS methods and search datasets which shows the relevance of augmentation strategy over other design choices.

\end{document}